\documentclass[10pt,letterpaper]{article}
\usepackage[top=0.85in,left=2.75in,footskip=0.75in]{geometry}

\usepackage[left,mathlines]{lineno}

\usepackage{cite}
\usepackage{amsmath,amssymb,amsfonts}
\usepackage{algorithm}
\usepackage{multirow}
\usepackage[]{graphicx}
\usepackage{algpseudocode} 
\usepackage{textcomp}
\usepackage{booktabs}
\usepackage{authblk}
\providecommand{\keywords}[1]{\textbf{Keywords:} #1}
\usepackage{bbm}
\usepackage{comment}
\usepackage{adjustbox}
\usepackage{makecell}
\usepackage{url}
\usepackage{bm}
\usepackage{rotating}
\usepackage{changepage}

\usepackage{textcomp,marvosym}

\usepackage{cite}
\usepackage{rotating}
\usepackage{nameref,hyperref}

\usepackage[nopatch=eqnum]{microtype}
\DisableLigatures[f]{encoding = *, family = * }

\usepackage[table]{xcolor}

\usepackage{array}

\newcolumntype{+}{!{\vrule width 2pt}}

\newlength\savedwidth

\raggedright
\usepackage[aboveskip=1pt,labelfont=bf,labelsep=period,justification=raggedright,singlelinecheck=off]{caption}

\makeatletter
\renewcommand{\@biblabel}[1]{\quad#1.}
\makeatother

\usepackage{lastpage,fancyhdr,graphicx}
\usepackage{epstopdf}
\fancyheadoffset[L]{2.25in}
\fancyfootoffset[L]{2.25in}
\begin{document}
\vspace*{0.2in}

\begin{flushleft}
{\Large
\textbf{Conditional Average Treatment Effect Distillation for Hypothetical State-Intervention Targeting Under Budget Constraints} 
}
\newline

Vasundhara Acharya\textsuperscript{1},
B\"{u}lent Yener\textsuperscript{2}

\bigskip
\textbf{1} Rensselaer Polytechnic Institute, Troy, USA (e-mail: acharv2@rpi.edu)
\\
\textbf{2} Rensselaer Polytechnic Institute, Troy, USA (e-mail: yener@cs.rpi.edu)
\\

\bigskip

%
%





\end{flushleft}


%

\section*{Abstract}
Personalized public-health targeting requires methods that capture heterogeneous intervention benefit while remaining interpretable under resource constraints. Flexible causal-effect models can be difficult to translate into operational rules, whereas coarse subgroup policies may lose individualized treatment-effect information. We study budget-constrained hypothetical state interventions in observational data without a logged historical decision policy. For obesity, elevated glucose, and lifetime smoking-history states, causal-structure-informed pretreatment covariates are prespecified before policy learning, and intervention capacity is capped at 70\% of individuals observed in the adverse exposure state. We evaluate a causal distillation framework in which causal forest and Bayesian Causal Forest (BCF) teachers estimate conditional average treatment effects (CATEs), and decision tree (DT) and random forest (RF) students approximate the resulting benefit surfaces. Individualized policies are compared with an unsupervised K-means subgroup policy and a supervised causal-tree policy with leaf count matched to the number of K-means clusters. All observational policies are evaluated on held-out data using a doubly robust estimator, bootstrap uncertainty, and decision-focused teacher--student fidelity measures. Across the PIMA obesity and glucose experiments and the NHANES smoking-history experiment, every distilled student was within 0.011 absolute utility of its corresponding teacher, with allocation Jaccard similarities of 0.738--0.937. K-means and causal-tree policies were also competitive, and the highest point-estimate policy varied across applications. Prespecified paired comparisons were not significant after Holm adjustment, while agreement near the budget boundary was less stable than global fidelity in some teacher--student pairs. Across 10 semi-synthetic IHDP realizations with known potential outcomes, BCF showed the strongest teacher performance. These results support causal distillation as a practical approach for translating flexible causal-effect models into simpler budgeted targeting rules, while emphasizing that policy value, allocation agreement, and boundary fidelity should be assessed alongside predictive fit. Observational estimates remain contingent on causal-identification assumptions and require longitudinal or prospective validation.

\keywords{Causal inference; Heterogeneous treatment effects; Causal distillation; Policy learning}


%


\section{Introduction}
\label{intro_causal}

Many clinical and public-health interventions are guided by individualized treatment rules that recommend treatment as a function of a patient’s pre-treatment covariates \cite{murphy}\cite{murphy3}\cite{imai}. A standard causal framework formalizes this through the conditional average treatment effect (CATE) and seeks policies that treat individuals with large, positive CATEs \cite{athey_policy}. Modern machine-learning methods such as causal forests and meta-learners  \cite{causalforest}\cite{chernozhukov2018double} can flexibly estimate heterogeneous treatment effects from observational data, i.e., individual-specific variation in expected intervention benefit driven by pre-treatment covariates \cite{HTE1}\cite{HTE2}. Still, the resulting models are often complex, difficult to interpret, and expensive to deploy.

In supervised prediction, knowledge distillation compresses a large teacher model into a smaller student by training the student to mimic the teacher’s outputs. We adapt this idea to the causal setting and study causal distillation, in which we compress a high-capacity CATE estimator into simpler student models and their corresponding treatment policies. Here, the teachers are causal forest \cite{causalforest} and Bayesian causal forest (BCF) \cite{hahn2020bcf} models fitted to observational data, and their outputs are individualized CATE estimates. The students are simple tree-based models trained to regress on the teacher’s CATEs and then define treatment policies by thresholding predicted benefits. We focus on clinically interpretable state interventions targeting modifiable risk-factor states (e.g., obesity, elevated glucose), where the observed exposure indicates baseline risk-factor status and the targeting rules are ranked by the construct.

Causal distillation is substantially more delicate than standard
distillation for predictive risk.  First, the ground-truth individual
treatment effect is never observed in an observational dataset \cite{alaa2018limit}, so we cannot directly
measure the accuracy of either the teacher or the students.  Second, in
practice we care about the policy induced by a model which deals with who is
treated and with what expected outcome rather than the CATE surface
itself.

In this study, we address the following questions:

\begin{itemize}
    \item How do unsupervised and supervised subgroup policies compare with individualized causal policies?
\item Can DT/RF students preserve teacher policy value and allocation behavior while reducing complexity?
\item Do these conclusions persist when true treatment effects and oracle policies are available in IHDP?
\end{itemize}

To address these questions, we evaluate a set of subgroup-level and
individualized targeting policies motivated by hypothetical state
interventions. For the observational-data experiments, intervention
eligibility, adjustment variables, and effect modifiers are specified from
the assumed causal structure before policy learning rather than being learned
as part of the proposed framework. Because the observed data arise from
natural exposure assignments rather than a deployed historical decision
policy, we evaluate the resulting hypothetical state-intervention policies
using held-out doubly robust estimation under standard causal-identification
assumptions. This work makes the following contributions:

\begin{itemize}

    \item We construct K-means subgroups using pretreatment covariates alone and
    subsequently rank these groups using estimated treatment benefit under a
    common intervention budget. We compare this unsupervised subgroup
    construction with a supervised causal-tree comparator matched in subgroup
    complexity, as well as with individualized policies learned from causal
    forests and BCF.

    \item 
    We distill treatment-effect estimates from both causal-forest and
    Bayesian-causal-forest teachers into simpler decision-tree and
    random-forest students. Beyond predictive fidelity, we evaluate whether
    the students preserve the teachers' treatment-effect rankings,
    budget-constrained allocation decisions, and held-out policy utility,
    thereby quantifying the tradeoff between structural simplicity and policy
    fidelity.

    \item 
    We evaluate teacher, student, and subgroup policies using a common
    selective-shift doubly robust estimator on held-out data. This permits
    direct comparison of policies that differ substantially in how they
    represent heterogeneity from coarse pretreatment-defined subgroups to
    individualized treatment-effect rankings while maintaining the same
    intervention eligibility and budget.

    \item 
    Using multiple Infant Health and Development Program (IHDP) realizations with known potential outcomes, we
    complement the observational-data analysis with evaluation based on precision in the estimation of heterogeneous effects (PEHE), true policy value, policy regret, treatment-effect
    ranking agreement, and overlap with the oracle top-$q$ policy. We further
    show theoretically in the Supplementary Material why treatment-effect
    estimation error and policy performance provide complementary information,
    relating policy regret to both CATE-estimation error and disagreement with
    the oracle policy.

\end{itemize}

 The term ground-truth individual treatment effect (ITE) specifically refers to oracle individual effects available only in synthetic or randomized benchmarks. In the observational setting, the ITE reported by teachers or students corresponds to a CATE estimate.

The remainder of this paper is organized as follows. Section~\ref{related} reviews related work. Section~\ref{methodology} presents the proposed workflow. Section~\ref{setups} describes the experimental setup. Section~\ref{Results} reports and analyzes the results. Section~\ref{discussion_causal} revisits the research questions introduced in Section~\ref{intro_causal}. Section~\ref{limit_future} discusses limitations and future research directions. Finally, Section \ref{conclusion_causal} concludes the paper.

\section{Related Work} \label{related}

\par Estimating how the causal effect of an intervention varies across
individuals or subgroups is central to personalized medicine and targeted
policymaking. Flexible heterogeneous treatment-effect estimators can capture
complex variation in treatment response, but state-of-the-art approaches
often rely on ensemble methods or deep learning models, both of which present substantial challenges to interpretability.
This situation creates a trade-off between the flexibility required for accurate conditional average treatment effect (CATE)
estimation and the transparency required for treatment targeting and
clinical decision-making.

Subgroup identification for heterogeneous treatment effects has been studied
using several supervised and effect-informed approaches. Foster et al.
\cite{virtualtwins} introduced the Virtual Twins procedure, which first
predicts each individual's outcome under treatment and control and then
models the resulting treatment contrast to identify subgroups with enhanced
effects. Athey and Imbens \cite{HTE_P3} proposed causal trees, which
recursively partition the pretreatment covariate space to identify regions
with differing treatment effects. Causal forests extend this approach
through ensembles of honest trees, providing flexible, nonparametric CATE
estimates \cite{causalforest}. Chernozhukov et al. \cite{GATES} instead
form groups by ranking individuals according to a machine-learning proxy for
treatment benefit and then estimate group-level effects using sample
splitting. Related effect-informed clustering methods include Causal
K-Means, which clusters individuals according to estimated counterfactual
response functions rather than similarity in observed baseline covariates
\cite{causalkmeans}.

Although these methods can identify treatment-effect heterogeneity, their
resulting models or subgroup structures often remain difficult to interpret
or deploy. Causal distillation addresses this limitation by transferring the
estimated treatment-effect structure of a flexible teacher into a simpler
student model with the goal of preserving treatment ranking and policy
performance. In this context, causal distillation links flexible heterogeneous treatment effect (HTE)
estimation with the construction of auditable treatment-assignment rules.

Makar et al. \cite{paper10_cd} introduced a two-stage distillation approach
for individual treatment-effect estimation. A rich teacher model first used
a broad set of covariates to adjust for confounding and estimate
heterogeneous effects. A student model was then trained using a smaller set
of features sufficient for predicting the estimated ITE. This approach
separated ITE discovery, which may require many covariates, from ITE
prediction for new individuals, which may depend primarily on a smaller set
of effect modifiers. By distilling the complex model into a simpler model
based on selected features, DEITEE achieved accurate ITE estimates with
reduced data requirements, although the approach relied on the assumption
that the relevant effect modifiers could be identified.

Rehill et al. \cite{paper9_cd} proposed a distilled causal tree for
interpretable HTE estimation. Their method first fitted a causal-forest
teacher to obtain CATE predictions and then trained a single decision-tree
student using a regression-to-teacher objective. To preserve causal
validity, the distilled tree was treated as an adaptively selected
partition, and leaf-level treatment effects were re-estimated on held-out
data using a doubly robust augmented inverse probability weighting (AIPW) estimator. The authors also examined
tree-search procedures intended to improve stability and fidelity, and
reported strong performance in noisy and high-dimensional settings.

Huang, Tang, and Kenney \cite{CDT1} introduced Causal Distillation Trees,
a two-stage approach for obtaining interpretable subgroups from black-box
ITE models. A flexible teacher was first used to estimate individual
treatment effects, after which a decision-tree student was trained on the
estimated effects to recover meaningful treatment-effect subgroups. Their
objective was not merely to improve predictive accuracy, but to improve the
theoretical consistency for the recovered subgroups and demonstrated the
approach using clinical-trial data.

Related uses of causal distillation have also appeared in adjacent
application areas. Sun et al. \cite{kdsm} proposed KDSM for uplift modeling
in marketing, in which an interpretable uplift-tree teacher transfers
incremental-outcome information to a neural-network student. The method also
uses counterfactual sample matching within tree leaves to augment training
data. Liu et al. \cite{recommend} introduced KDCRec for recommender systems,
where a teacher trained on less biased data guides a student trained on
biased observational data. More broadly, causal concepts have been
incorporated into knowledge distillation for language modeling, bias
mitigation, and continual learning
\cite{paper1_cd,paper2_cd,paper3_cd,paper7_cd}. These applications
demonstrate the broad applicability of causal distillation.


\par Once heterogeneous treatment effects have been estimated and distilled
into a treatment rule, a separate question concerns how the resulting rule
should be evaluated in practice.

Evaluating individual treatment-effect estimators is challenging in
observational data because the true individual treatment effects are
unobserved. In semi-synthetic or benchmark settings with known
counterfactual outcomes, estimators can be evaluated using measures such as
mean squared error, precision in the estimation of heterogeneous effects
(PEHE), and error in the average treatment effect (ATE). For example, Shalit
et al. \cite{shalit} used PEHE and (ATE) error on the Infant Health and
Development Program benchmark. When ground-truth individual effects are
unavailable, policy-based evaluation provides an alternative by assessing
the expected outcome or utility induced by treatment decisions based on the
the estimated effects for the target population.

\par Policy value may be estimated using outcome-regression, inverse-propensity-weighted, or doubly robust estimators. Doubly robust evaluation combines outcome-model predictions with a propensity-weighted residual correction and remains consistent when either the outcome model or the propensity model is correctly specified. Dudík
et al. \cite{dudik2011doubly} established doubly robust methods for policy
evaluation and learning, while Sverdrup et al. \cite{athey} combined
doubly robust scores with sample splitting and cross-fitting for policy
learning using observational data.

A related, but distinct, literature directly optimizes treatment-assignment
rules rather than first estimating and distilling individual effects.
Kitagawa and Tetenov \cite{kitagawa2018treated} proposed empirical welfare
maximization over constrained policy classes, which is particularly relevant
when treatment rules must remain interpretable or are subject to resource
constraints. In multi-action settings, Zhou et al. \cite{zhou} constructed
cross-fitted doubly robust scores and optimized them over interpretable
estimated welfare. In contrast, our study first estimates heterogeneous
treatment effects and then evaluates whether the resulting information can be
transferred from a flexible teacher model to an interpretable student model while
preserving held-out policy value.

\par The reliability of the final treatment rule also depends on decisions
made before HTE estimation. In observational data, one such decision is the
selection of covariates used for confounding adjustment. Because the
adjustment set in our framework is informed by a discovered causal graph,
the stability of causal discovery is therefore relevant to the overall
analytical pipeline.
Ensemble learning strategies are increasingly employed in causal discovery
to mitigate the instability, overfitting, and hyperparameter sensitivity
often exhibited by individual structure-learning algorithms
\cite{paper1_causal_disc,paper3_causal_disc,paper5_causal_disc,
paper10_causal_disc}. By aggregating results across multiple data
partitions, bootstrap samples, or diverse algorithmic outputs, these
frameworks aim to reduce false-positive edges and improve the stability of
the inferred causal graph
\cite{paper2_causal_disc,paper4_causal_disc,paper6_causal_disc,
paper8_causal_disc,paper9_causal_disc,paper11_causal_disc}.

Taken together, these bodies of literature provide flexible HTE estimators,
interpretable distillation procedures, policy-evaluation methods, and tools
for improving causal-discovery stability. However, comparatively limited
work integrates these components into an end-to-end framework for
observational public health targeting.

Prior causal-distillation methods commonly assume interventional
supervision, such as randomized-trial data, or rely on a fixed,
analyst-specified adjustment set. They are also frequently evaluated using
ground-truth ITE metrics on synthetic or semi-synthetic benchmarks
\cite{paper9_cd,Policy_risk1}. In contrast, observational public-health
applications require confounding adjustment using an estimated treatment
mechanism, may involve complex survey designs, and often lack a historical
logged decision policy \cite{austin}. Nevertheless, they require
actionable and auditable treatment rules over modifiable exposure states,
such as obesity or elevated glucose
\cite{DT_intrepretable2}.

There remains limited end-to-end work that combines data-informed
identification of adjustment variables, confounding-adjusted HTE
estimation, interpretable distillation, and held-out policy evaluation in
this setting. Existing studies also rarely evaluate distillation using
decision-focused criteria such as treatment agreement, ranking stability,
and policy value, particularly near an intervention threshold where small
approximation errors can change treatment assignment. Our framework
addresses these gaps by integrating covariate selection informed by causal structure,
selection, flexible teacher estimation, interpretable student models, and
held-out doubly robust policy evaluation.

\section{Methodology}\label{methodology}

The proposed framework is evaluated across observational and semi-synthetic datasets using separate discovery and evaluation cohorts. Model fitting, subgroup construction, and policy formation are conducted on the discovery data, after which the resulting policies are fixed and assessed on the held-out evaluation cohort. Observational policies are evaluated using doubly robust estimation, while the semi-synthetic benchmark also allows for comparison with known counterfactual outcomes and oracle policies. The following subsections outline the datasets, policy construction, causal estimators, subgroup comparators, distillation procedure, and evaluation criteria.

\subsection{Datasets}

This study utilizes the PIMA Indians Diabetes Dataset
\cite{PIMA_diabetes_dataset}, sourced from the National Institute of Diabetes
and Digestive and Kidney Diseases, and the National Health and Nutrition
Examination Survey (NHANES) 2015--2020 pre-pandemic data
\cite{nhanes_data,Guidelines_nhanes}. The PIMA Indians Diabetes dataset is
publicly available at
\url{https://www.kaggle.com/datasets/mdssh784/pima-indians-diabetes-dataset},
and the NHANES public-use data files are available at
\url{https://wwwn.cdc.gov/nchs/nhanes/}. Detailed descriptions of the
datasets, variables, and analytic cohorts are provided in the Supplementary
Material. We also present our results using the 10 publicly available NPCI realizations from the CEVAE
repository~\cite{louizos2017causal}, which have been adopted at the same
scale in prior work~\cite{rovai2026civex,vo2025federated,ashman2023causal}.

To avoid introducing additional assumptions associated with simple
imputation procedures such as mean or median imputation, observations with
missing values in variables required for a given analysis were excluded,
resulting in complete-case analytic cohorts \cite{acharya2026unsupervised}. For the NHANES smoking-history policy, to obtain a continuous Euclidean representation suitable for K-means clustering, these variables were one-hot encoded and transformed
using Multiple Correspondence Analysis (MCA) \cite{greenacre1984theory,greenacre2006multiple,abdi2007multiple}. For the NHANES smoking-history subgroup analysis, the first seven MCA dimensions were retained as a fixed lower-dimensional representation for downstream clustering. NHANES survey weights were not incorporated in the present policy-learning
analysis. Accordingly, the reported NHANES results characterize the
constructed analytic sample and should not be interpreted as
survey-weighted, nationally representative estimates of the U.S. population.

\subsubsection{Data access and identifiability}

This study is a secondary analysis of publicly available, de-identified
datasets. The observational analyses used the NHANES 2015--2020
pre-pandemic public-use files and the PIMA Indians Diabetes Dataset.
The NHANES data were accessed for research purposes on 21 August 2024,
and the PIMA Indians Diabetes Dataset was accessed on 15 July 2025. We additionally used the publicly available semi-synthetic Infant Health
and Development Program (IHDP) benchmark, specifically the 10 NPCI
realizations distributed through the CEVAE repository. The IHDP benchmark
combines observed pretreatment covariates with simulated potential outcomes,
thereby providing known counterfactual outcomes and treatment effects for
methodological evaluation. The IHDP data was accessed on
15th May, 2026. All datasets used in this study were publicly available and contained no direct participant identifiers accessible to the authors. The authors did not have access to information that could identify individual participants at any time during or after the analyses.

\subsubsection{Causal-structure-informed covariate specification}
\label{subsubsec:causal_covariates}

Candidate pretreatment covariates, adjustment variables, and effect
modifiers were specified using the three-stage ensemble causal-structure
discovery procedure developed in our previous work
\cite{acharya2025understanding}. Briefly, the procedure combines multiple
causal-discovery algorithms with adaptive bootstrapping to estimate stable
edge-confidence scores, resolve directional conflicts and cycles, and
aggregate algorithm-specific graphs into a consensus directed acyclic
graph. The resulting structure was further reviewed using clinical
expertise and evidence from the epidemiologic literature.

For PIMA, the ensemble incorporated Fast Causal Inference (FCI) and Peter--Clark (PC) with kernel
conditional-independence testing, nonlinear NOTEARS, Pillai's test, and
supplementary additive-noise-model direction tests
\cite{spirtes2001anytime,zheng2020learning,pillai_test_ankur,
hoyer2008nonlinear}. For NHANES, the ensemble combined PC with discrete
conditional-independence tests, greedy hill climbing under multiple
discrete-data scores, and Fast Greedy Equivalence Search (FGES) with the Bayesian Dirichlet equivalent uniform score (BDeu) score
\cite{PC,hill_climb,fast_ges}. Full algorithmic details, bootstrap
procedures, confidence aggregation, and graph-validation analyses are
reported in \cite{acharya2025understanding}.

In the present study, the resulting causal structures are treated as
prespecified design-stage inputs rather than as outputs of the distillation
framework. They are used only to inform the selection and roles of
pretreatment covariates for each policy experiment. Causal discovery is not
repeated within individual discovery--evaluation splits. The prespecified
adjustment set is held fixed for teacher estimation and policy evaluation,
while the designated deployable pretreatment features are supplied to the
student models and subgroup-policy comparators.

Because the graph is treated as a fixed design input, uncertainty in its
structure is not propagated through the downstream evaluations.
Possible graph misspecification may therefore affect the selected covariate roles and remains a limitation of the analysis.


\subsection{Budgeted Selective State-Shift Policies}
\label{subsec:state_shift_estimand}

We evaluate hypothetical interventions on modifiable or clinically
interpretable exposure states rather than treatments assigned by a historical
decision rule. The formal intervention-specific doubly robust policy-value
estimator used here was developed in our previous work
\cite{acharya2026unsupervised}, building on
framework of Wen et al.~\cite{wen2023intervention}. Here, we summarize the
policy estimand and specify its application to the body mass index (BMI), glucose, and
smoking-history experiments.

Let $T_i\in\{0,1\}$ denote the experiment-specific exposure state, where
$T_i=1$ represents the adverse state and defines eligibility for the
hypothetical intervention, and let $T_i=0$ denote the corresponding
lower-risk state. Let $X_i$ denote the pre-treatment variables, $L_i$ the pre-treatment adjustment variables used for
causal evaluation, and $Y_i(t)$ the potential outcome under exposure state
$t$. In the present analyses, $Y_i=1$ denotes an adverse binary outcome. In the estimator and algorithm notation below, we use $Z_i \equiv L_i$ to denote this same adjustment-variable set.

A learned policy $\pi$ induces a shift probability
$s_{\pi}(X_i)\in[0,1]$. The intervention is activated only for individuals
observed in the adverse state. Specifically,

\begin{equation}
T_i^{\pi}
=
T_i\left(1-S_i^{\pi}\right),
\qquad
S_i^{\pi}\sim
\operatorname{Bernoulli}\!\left(s_{\pi}(X_i)\right)
\quad\text{only when }T_i=1.
\label{eq:selective_state_shift}
\end{equation}

Thus, a prioritized eligible individual may be shifted from
$T_i=1$ to $T_i=0$, whereas individuals already observed with $T_i=0$
remain in their factual state. For deterministic individualized policies,
$s_{\pi}(X_i)$ is generally zero or one. For group-based policies, a
fractional value may arise when the intervention budget ends within a
boundary subgroup.

The intervention budget is defined relative to the eligible population.
For an evaluation cohort containing

\begin{equation}
N_{\mathrm{elig}}=\sum_i \mathbbm{1}\{T_i=1\},
\end{equation}

the maximum intervention capacity is

\begin{equation}
B_q=\operatorname{round}\left(qN_{\mathrm{elig}}\right),
\label{eq:eligible_budget}
\end{equation}

where $q\in(0,1]$ denotes the prespecified budget fraction. The budget is a
maximum capacity rather than a compulsory quota. Individuals or subgroups
with nonpositive estimated benefit are not selected merely to exhaust the
available budget.

For a candidate policy $\pi$, the policy-risk estimand is the expected
adverse outcome under the induced selective state shift,

\begin{equation}
R(\pi)
=
\mathbb{E}\!\left[Y_i^{\pi}\right]
=
\mathbb{E}\!\left[Y_i\!\left(T_i^{\pi}\right)\right].
\label{eq:state_shift_risk}
\end{equation}

Importantly, eligibility determines who may be shifted. It does not
restrict policy-value evaluation to individuals with $T_i=1$. Individuals
already observed in the lower-risk state remain part of the evaluation
population and contribute information about the target-state outcome
distribution.

Under consistency, no interference, conditional exchangeability given
$L_i$, and positivity, the selective state-shift risk is identified from the
observational data. As in our previous work
\cite{acharya2026unsupervised}, we estimate policy value on the held-out
evaluation cohort using an intervention-specific doubly robust score. To reduce instability from extreme
propensity scores, evaluation is restricted to observations satisfying

\begin{equation}
A_i
=
\mathbbm{1}
\left\{
\alpha
\leq
\widehat e(L_i)
\leq
1-\alpha
\right\},
\label{eq:state_shift_overlap}
\end{equation}

where $e(L_i)=P(T_i=1\mid L_i)$ and $\alpha=0.05$ in the primary analyses.
Accordingly, the reported policy-risk estimand is

\begin{equation}
R_A(\pi)
=
\mathbb{E}
\left[
Y_i^{\pi}\mid A_i=1
\right],
\label{eq:trimmed_policy_risk}
\end{equation}

and, because the outcomes considered here are adverse binary outcomes,
policy utility is defined as

\begin{equation}
U_A(\pi)
=
1-R_A(\pi).
\label{eq:state_shift_utility}
\end{equation}

All individualized teacher policies, distilled-student policies, and
group-based comparators are evaluated under this same held-out estimand and
doubly robust evaluation procedure.

\subsection{Hypothetical State-Shift Policies}
\label{subsec:pima_state_policies}

For the PIMA dataset, we consider two policy experiments involving modifiable metabolic
risk-factor states: obesity and elevated glucose. These experiments are
motivated by randomized diabetes-prevention evidence demonstrating that
weight and glycemic status are modifiable and strongly associated with
subsequent diabetes risk. In the Diabetes Prevention Program, intensive
lifestyle intervention targeting weight reduction and physical activity
substantially reduced incident type 2 diabetes \cite{proof_bmi}, with weight
loss emerging as an important predictor of risk reduction
\cite{proof_bmi2}. Interventions targeting impaired glucose tolerance have
similarly produced sustained reductions in diabetes incidence
\cite{proof_glucose1}, and regression from prediabetes to normal glucose
regulation has been associated with substantially lower subsequent diabetes
risk \cite{proof_glucose2}. These findings motivate the state contrasts
considered here. 
For the NHANES experiment, the adverse-state indicator is based on reported
lifetime smoking history. However, the evaluated policies should nevertheless be interpreted
as hypothetical state interventions rather than specific clinical treatment
regimens.

\paragraph{BMI policy.}
For the BMI experiment, the adverse-state indicator is

\begin{equation}
T_i:=T_{\mathrm{BMI},i}=\mathbbm{1}
\left\{
\mathrm{BMI}_i
\geq
30~\mathrm{kg/m^2}
\right\}.
\label{eq:bmi_state_indicator_causal}
\end{equation}

Thus, $T_i=1$ denotes obesity and defines eligibility for the hypothetical
$1\rightarrow0$ state shift, whereas $T_i=0$ denotes the non-obese state.
The threshold follows the standard adult definition of obesity
\cite{purnell2015definitions}.

\paragraph{Glucose policy.}
For the glucose experiment, we define

\begin{equation}
T_i:=T_{\mathrm{Glucose},i}=\mathbbm{1}
\left\{
\mathrm{Glucose}_i
\geq
140~\mathrm{mg/dL}
\right\}.
\label{eq:glucose_state_indicator_causal}
\end{equation}

Here, $T_i=1$ denotes the elevated-glucose state and defines eligibility for
the hypothetical shift, whereas $T_i=0$ denotes the lower-glucose state.
The $140~\mathrm{mg/dL}$ threshold corresponds to the lower boundary of
impaired glucose tolerance for the two-hour oral glucose tolerance test
\cite{bansal2015prediabetes}.

For both PIMA experiments, the outcome is diabetes status, with $Y_i=1$
denoting diabetes. The intervention budget is applied only to individuals
observed in the corresponding adverse state, while policy risk and utility
are evaluated over the overlap-restricted held-out cohort according to
Eqs.~\eqref{eq:trimmed_policy_risk} and
\eqref{eq:state_shift_utility}.

\paragraph{Smoking policy.}
For the NHANES Smoking experiment, we define

\begin{equation}
T_i:=T_{\mathrm{Smoking},i}=\mathbbm{1}
\left\{
\text{reported smoking at least 100 cigarettes during the lifetime}
\right\}.
\label{eq:smoking_state_indicator}
\end{equation}

Thus, $T_i=1$ denotes the lifetime-smoking-history state and defines
eligibility for policy allocation, whereas $T_i=0$ denotes the corresponding
non-smoking-history state. Because this exposure captures lifetime smoking
history rather than current smoking status, the hypothetical
$1\rightarrow0$ contrast should not be interpreted as an acute
smoking-cessation intervention. Instead, it represents a counterfactual
contrast between the established lifetime-smoking and non-smoking-history
states among prioritized eligible individuals.

The outcome is sleep disturbance, with $Y_i=1$ denoting the adverse outcome.
The intervention budget is again defined relative to the eligible
$T_i=1$ population, while both adverse- and lower-risk-state individuals
contribute to held-out policy-value estimation under the common selective
state-shift estimand in Eq.~\eqref{eq:trimmed_policy_risk}.

\subsection{Group-Based Policy Comparators}
\label{subsec:group_policy_comparators}

To compare the individualized teacher--student policies with simpler
group-based decision rules, we consider two subgroup-policy comparators:
unsupervised K-means clustering and a supervised Causal Tree. Both methods
use the same downstream gain-based ranking, intervention budget, and held-out
policy-evaluation procedure. They differ primarily in how the subgroup
structure is learned.

\paragraph{Unsupervised K-means subgroup policy}

K-means clustering is performed using only prespecified pre-treatment
covariates, without treatment, outcome, or estimated treatment-effect
information. Thus, treatment-effect information is introduced only after the
clusters have been formed. This design is intended to represent a simple
phenotype-based unsupervised comparator rather than a causal HTE estimator.
Such unsupervised clustering approaches have been used to identify clinically
relevant patient phenotypes and subgroups, including in oncology
\cite{xu2023unsupervised}, and have also been investigated specifically for
assessing heterogeneity of treatment effects across baseline patient
phenotypes \cite{bellavia2025unsupervised}. Accordingly, K-means provides a
clinically motivated benchmark for assessing whether effect-informed
individualized or supervised subgroup policies offer additional value beyond
grouping patients according to pretreatment characteristics alone.

Clustering variables are standardized using discovery-cohort means and
standard deviations, with the fitted transformation subsequently applied to
the evaluation cohort \cite{scaling_clustering2,scaling_clustering}. The
number of clusters is selected using the silhouette score
\cite{silhoutte_score_paper}, evaluating candidate values
$K=2,\ldots,9$ on the discovery cohort only. The complete silhouette-score
profile is provided in the supplementary files.

After fixing the K-means partition, treatment-stratified out-of-fold (OOF) outcome
models are used to obtain discovery-cohort benefit scores as shown in the Eq.\ref{eq:kmeans_individual_benefit}.

\begin{equation}
\widehat{\tau}_i
=
\widehat{\mu}_1(Z_i)-\widehat{\mu}_0(Z_i),
\label{eq:kmeans_individual_benefit}
\end{equation}

where positive values indicate estimated benefit from the one-way shift
$T=1\rightarrow T=0$ for the adverse binary outcomes considered here.

\paragraph{Supervised Causal Tree.}

As a supervised subgroup comparator, we additionally fit a Causal Tree \cite{HTE_P3}. Unlike K-means, the tree uses treatment and outcome information to construct subgroups with heterogeneous treatment effects. To provide comparable subgroup complexity, the maximum number of terminal leaves is fixed to the same K selected for K-means, with each terminal leaf treated as a subgroup. The discovery cohort is internally split for tree-structure learning and honest leaf-effect estimation using cross-fitted AIPW effect scores. After fixing the partition, mean leaf benefit is estimated among eligible individuals, and leaves are ranked by aggregate discovery gain. Only positive-gain leaves are eligible for allocation, and the resulting partition and ranking are frozen before held-out evaluation.

\paragraph{Common subgroup policy.}

For either method, let $g$ denote a K-means cluster or Causal Tree leaf.
The individual benefit score is method-specific: K-means uses the
treatment-stratified out-of-fold outcome-model contrast in
Eq.~\ref{eq:kmeans_individual_benefit}, whereas the Causal Tree uses the
cross-fitted AIPW effect score from the honest leaf-estimation sample.
Among eligible discovery-cohort individuals, these scores are
averaged within subgroup $g$ to obtain the subgroup mean benefit
$\widehat{\tau}_g$ in Eq.~\ref{eq:group_mean_benefit}.

\begin{equation}
\widehat{\tau}_g
=
\frac{1}{n_g}
\sum_{i:G_i=g,\;T_i=1}
\widehat{\tau}_i,
\label{eq:group_mean_benefit}
\end{equation}

and the aggregate gain is computed using the Eq. \ref{eq:group_gain}.

\begin{equation}
\widehat{G}_g
=
n_g\widehat{\tau}_g,
\label{eq:group_gain}
\end{equation}

where $n_g$ is the number of eligible discovery individuals in subgroup $g$.
Only subgroups with strictly positive gain are considered, and these are
ranked in descending order of $\widehat{G}_g$.

The subgroup ranking is frozen before evaluation. In the held-out evaluation
cohort, positively ranked subgroups are allocated sequentially up to the
maximum budget as shown in the Eq.\ref{eq:eligible_budget}.

In the present analysis, subgroup allocation is intentionally based on the estimated mean benefit and aggregate gain alone. The within-subgroup variance of the estimated benefits is not incorporated into the ranking or budget-allocation rule. Thus, the subgroup comparators considered here are point-estimate-based policies rather than uncertainty-gated policies. Extensions that explicitly incorporate subgroup-level estimation uncertainty, including an Empirical Bernstein safety gate and a fully Bayesian hierarchical pooling gate, are developed and evaluated separately in our related work \cite{acharya2026unsupervised}.

Subgroups that fit within the remaining budget are fully selected. If the
budget boundary falls within a subgroup, all members of that subgroup are
assigned the corresponding common fractional shift probability. Subgroups
below the boundary or with nonpositive discovery gain receive zero shift
probability. The intervention itself is active only for currently eligible
individuals ($T_i=1$), and $q$ is treated as a maximum capacity rather than
a compulsory quota.


\subsection{Teacher and Student Models}
\label{subsec:teacher_student_models}

Our distillation framework includes two teacher models and two student models. The teacher models are a causal forest and a BCF \cite{hahn2020bcf}, both of which estimate a continuous conditional average treatment effect (CATE) for each individual. The student models are a decision tree and a random forest trained to approximate the CATE surface estimated by the corresponding teacher. Thus, the same two student architectures are evaluated separately using targets generated by each teacher.

\subsubsection{Causal-Forest Teacher}

Causal forests extend Breiman's random forests to estimate heterogeneous treatment effects nonparametrically. Under standard unconfoundedness and regularity assumptions, causal forests yield point-consistent CATE estimates with asymptotically normal sampling distributions, thereby supporting the construction of confidence intervals for $\tau(x)$ \cite{causalforest}. By adaptively partitioning the covariate space, causal forests can capture nonlinear relationships and high-order interactions between covariates and treatment without requiring a prespecified parametric treatment-effect model. They have also been shown empirically to outperform classical approaches, such as nearest-neighbor matching, in settings containing irrelevant covariates \cite{causalforest}.

In our implementation, the causal forest is estimated using a double machine learning (DML) formulation with orthogonalization. The treatment and outcome nuisance functions are estimated separately, and the resulting residualization reduces the sensitivity of the treatment-effect estimates to misspecification of either nuisance model. Causal forests have been successfully applied in biomedical and clinical research to investigate treatment-effect heterogeneity \cite{causal_forest_app}, supporting their use as a flexible and well-studied teacher model in our distillation framework.

\subsubsection{Bayesian-Causal-Forest Teacher}

As a complementary teacher model, we use BCF \cite{hahn2020bcf}. BCF decomposes the conditional response surface into a prognostic component and a treatment-effect component.
This decomposition allows the treatment-effect surface to be regularized separately from the prognostic surface. Such separation is particularly useful in observational causal inference because treatment-effect heterogeneity may be weak relative to the overall variation in the outcome. Under the BCF parametrization, weak or noisy heterogeneity can be shrunk toward a more homogeneous treatment-effect surface, potentially producing more stable CATE estimates than response-surface models that do not directly control regularization of the treatment-effect component \cite{hahn2020bcf}.

Let $\tau_i^{(s)}$ denote the $s$-th posterior draw of the individual-level CATE for evaluation individual $i$, where $s=1,\ldots,S$. To maintain comparability with the main causal-forest distillation pipeline, in which each student is trained using a point-valued CATE target, we distill the BCF posterior mean CATE:
\begin{equation}
\bar{\tau}_i=\mathbb{E}[\tau_i\mid\mathcal{D}]
\approx
\frac{1}{S}\sum_{s=1}^{S}\tau_i^{(s)}.
\label{eq:bcf_post_mean_cate}
\end{equation}
This produces a single continuous target for each individual and therefore permits a direct comparison between the BCF- and causal-forest-based distillation pipelines.

\subsubsection{Student Models}

For each teacher model, we train two student models: a single decision tree (DT) \cite{loh2011classification} and a random forest (RF) \cite{breiman2001random}. Both students receive only the pretreatment covariates $X$ as inputs and are trained to regress on the CATE targets generated by the teacher. For the causal-forest teacher, the target is the estimated point-valued CATE, $\widehat{\tau}^{\mathrm{CF}}_i$. For the BCF teacher, the target is the posterior mean CATE, $\bar{\tau}^{\mathrm{BCF}}_i$, defined in Equation~\eqref{eq:bcf_post_mean_cate}.

Accordingly, the student models are not causal estimators in their own right. They do not directly model treatment assignment or observed outcomes, and they do not optimize a causal identification or policy-value objective. Instead, they should be interpreted as compact approximations to the treatment-effect surface learned by the corresponding causal teacher.

The decision-tree student provides a particularly transparent representation of the teacher through a collection of hierarchical if--then decision rules. The random-forest student can capture a more flexible approximation to the teacher's CATE surface while allowing variable-importance and tree-based summaries to be examined. Although the random forest is less directly interpretable than a single tree, both models offer substantially simpler predictive representations than the original causal teachers \cite{DT_interpretable,DT_intrepretable2}.

Algorithm~\ref{student_training_ITE} summarizes the proposed framework.
To characterize the learned heterogeneity in a clinically readable form, we
inspect representative decision regions from the distilled student models.
For the decision-tree student, terminal-node paths provide exact rules, and
we report one representative rule for each policy-status category: fully
prioritized, capacity-boundary, and positive-benefit but outside the current
intervention capacity. For the random-forest student, which does not admit a
single exact decision rule, we use shallow tree-based surrogate regions to
summarize the predictions of the full ensemble. These surrogate rules are
used only for interpretation and do not determine policy allocation.

\begin{algorithm*}[!h]
\caption{Teacher--Student Causal Policy Learning and Held-out Evaluation}
\label{student_training_ITE}
\begin{algorithmic}[1]

\State \textbf{Input:}
Discovery data
$\mathcal D_{\mathrm{disc}}=\{(X_i,Z_i,T_i,Y_i)\}$;
evaluation data
$\mathcal D_{\mathrm{eval}}$;
teacher models
$\mathcal M_T=\{\mathrm{CF},\mathrm{BCF}\}$;
student models
$\mathcal M_S=\{\mathrm{DT},\mathrm{RF}\}$;
budget cap $q$;
number of OOF folds $K_{OOF}$.

\State \textbf{Output:}
Teacher and student benefit scores, learned policies,
student--teacher fidelity measures, and held-out policy risk/utility.

\vspace{0.4em}

\For{each teacher $m \in \mathcal M_T$}

    \State Partition $\mathcal D_{\mathrm{disc}}$ into
    $K$ treatment-stratified folds.

    \State Initialize OOF teacher-benefit vector
    $\widehat{\boldsymbol{\tau}}^{(m)}_{\mathrm{OOF}}$.

    \For{$k=1,\ldots,K$}

        \State Fit teacher $m$ on
        $\mathcal D_{\mathrm{disc}}\setminus\mathcal I_k$
        using $(Y,T,X,Z)$ as required by the teacher model.

        \State Predict treatment-benefit estimates for observations in
        $\mathcal I_k$ and store them in
        $\widehat{\boldsymbol{\tau}}^{(m)}_{\mathrm{OOF}}$.

    \EndFor

    \State Fit the final teacher $m$ on the complete discovery cohort.

    \State Predict teacher benefit scores on the evaluation cohort:
    \[
        \widehat{\tau}^{(m)}_i,
        \qquad i\in\mathcal D_{\mathrm{eval}}.
    \]

    \For{each student $s \in \mathcal M_S$}

        \State Fit the non-causal student using only
        \[
        X_i
        \longrightarrow
        \widehat{\tau}^{(m)}_{\mathrm{OOF},i},
        \qquad i\in\mathcal D_{\mathrm{disc}}.
        \]

        \State Predict student benefit scores on the evaluation cohort:
        \[
        \widehat{\tau}^{(m,s)}_i
        =
        M_s(X_i).
        \]

    \EndFor

    \State Construct a policy for the teacher and each student by ranking
    currently eligible individuals ($T_i=1$) by predicted benefit.

    \State Retain only individuals with strictly positive predicted benefit
    and select up to the budget
    \[
        B_q =
        \operatorname{round}
        \left(
        q\,N_{\mathrm{elig}}
        \right).
    \]

    \State If multiple individuals share the budget-boundary score,
    assign the common fractional boundary shift probability required to
    satisfy the budget in expectation.

    \State Freeze all teacher and student policies before outcome evaluation.

    \State Estimate policy risk on
    $\mathcal D_{\mathrm{eval}}$ using the common cross-fitted
    selective one-way-shift doubly robust estimator with propensity
    adjustment and overlap trimming.

    \State For adverse binary outcomes, compute
    \[
        \widehat U(\pi)=1-\widehat R(\pi).
    \]

    \State Obtain $95\%$ bootstrap confidence intervals and paired
    teacher--student utility differences using common evaluation resamples.

    \State Compute student--teacher fidelity and policy-agreement measures
    on the evaluation cohort.

\EndFor

\State \Return teacher/student policies, held-out policy risk and utility,
confidence intervals, and fidelity measures.

\end{algorithmic}
\end{algorithm*}


\subsection{Metrics for Evaluation}
\label{sec:evaluation_metrics}

\begin{itemize}

\item \textbf{Policy risk and utility:}

For the observational PIMA and NHANES analyses, individual counterfactual
outcomes are not jointly observed, so treatment-effect accuracy cannot be
evaluated directly \cite{shalit}. We therefore evaluate the policies
induced by the estimated heterogeneous effects on the held-out evaluation
cohort, after all models and policy rules have been fixed using the discovery
cohort.

Under the selective state-shift estimand defined in
Section~\ref{subsec:state_shift_estimand}, policy risk is estimated using the
cross-fitted doubly robust estimator (DR)

\begin{equation}
\widehat{R}_{\mathrm{DR}}(\pi)
=
\frac{1}{n_{\mathrm{ov}}}
\sum_{i\in\mathcal{I}_{\mathrm{ov}}}
\left[
(1-T_i s_i)Y_i
+
T_i s_i \widehat{\mu}_0(Z_i)
+
(1-T_i)s_i
\frac{\widehat e(Z_i)}
     {1-\widehat e(Z_i)}
\left\{
Y_i-\widehat{\mu}_0(Z_i)
\right\}
\right],
\label{eq:selective_shift_dr}
\end{equation}

where $s_i=s_{\pi}(X_i)$,
$\widehat e(Z_i)=P(T_i=1\mid Z_i)$ is the estimated propensity score,
$\widehat{\mu}_0(Z_i)=E[Y_i\mid T_i=0,Z_i]$ is the outcome regression under
the target state, and $\mathcal{I}_{\mathrm{ov}}$ denotes the
overlap-restricted evaluation sample. Nuisance functions are estimated using
five-fold treatment-stratified cross-fitting. The estimator is doubly robust
under the standard identification conditions \cite{policy_risk2}. 

Because $Y=1$ denotes an adverse outcome, policy utility is shown as in the equation \ref{eq:policy_utility_causal}.

\begin{equation}
U(\pi)=1-\widehat{R}_{\mathrm{DR}}(\pi).
\label{eq:policy_utility_causal}
\end{equation}

We additionally report the no-shift and shift-all-eligible reference
policies. Learned policies are evaluated primarily under the prespecified
$q=0.70$ maximum eligible-population intervention budget.

\item \textbf{Policy agreement:}

For teacher--student comparisons, let $S_T$ and $S_S$ denote the eligible
individuals selected by the teacher and student, respectively, at the
prespecified budget. Selected-set agreement is measured using the Jaccard
index,

\begin{equation}
J(S_T,S_S)
=
\frac{|S_T\cap S_S|}
     {|S_T\cup S_S|}.
\label{eq:jaccard}
\end{equation}

The same measure is used when comparing selected populations from other
policies.

\item \textbf{Teacher--student fidelity:}

We assess how closely each student reproduces the heterogeneous-effect
signal estimated by its corresponding teacher. Prediction fidelity is
measured using the coefficient of determination ($R^2$) between the
teacher and student estimated benefit scores.

Because policy allocation depends on the ordering of estimated benefit, we
also assess ranking fidelity using Spearman's rank correlation coefficient
\cite{spearman3} and Kendall's $\tau_b$
\cite{spearman1,kendall2}. To evaluate agreement near the operational
allocation threshold, we additionally compute Kendall's $\tau_b$ within a
prespecified $\pm0.05$ quantile neighborhood around the teacher's budget
boundary.

\item \textbf{Paired comparison of policy utilities:}

Held-out utility differences are evaluated using a paired nonparametric
bootstrap. For policy $\pi_m$ and reference policy $\pi_{\mathrm{ref}}$,

\begin{equation}
\Delta U_{m,\mathrm{ref}}
=
U(\pi_m)-U(\pi_{\mathrm{ref}}).
\label{eq:paired_utility_difference}
\end{equation}

Evaluation individuals are resampled jointly for all policies within each
bootstrap replicate, preserving the pairing of policy values. The
corresponding teacher serves as the reference for teacher--student
comparisons, whereas K-means serves as the reference for subgroup-policy
comparisons. Because eight comparisons were prespecified within each
observational policy experiment, Holm's procedure
\cite{holm1979simple} is applied to the paired-bootstrap $p$-values to
control the family-wise error rate.

\item \textbf{Ground-truth metrics for semi-synthetic experiments:}

For IHDP, where simulated potential outcomes and true treatment effects are
available, we additionally report true policy value, policy regret,
agreement with the oracle budget-constrained policy, and treatment-effect
estimation accuracy using PEHE.

\item \textbf{Model complexity:}

Structural complexity is summarized by terminal-leaf count. For a single
decision tree, complexity is its number of terminal leaves. For a random
forest or causal forest containing $B$ trees,

\begin{equation}
C_{\mathrm{forest}}
=
\sum_{b=1}^{B} L_b,
\label{eq:forest_complexity}
\end{equation}

where $L_b$ is the number of terminal leaves in tree $b$
\cite{therneau1997introduction}.

For BCF \cite{hahn2020bcf}, we measure complexity using the treatment-effect
forest because this is the component distilled by the student models. For
posterior draw $m$,

\begin{equation}
C_{\tau}^{(m)}
=
\sum_{b=1}^{B_{\tau}}
L_{\tau b}^{(m)},
\label{eq:bcf_complexity}
\end{equation}

where $B_{\tau}$ is the number of treatment-effect trees and
$L_{\tau b}^{(m)}$ is the number of terminal leaves in tree $b$. We summarize
this quantity across retained posterior draws using its posterior mean and
interval.

\end{itemize}

\subsection{Ablation Studies}

\subsubsection{Sensitivity Analysis: Nuisance-model replacement}

To examine whether the estimated policy values were sensitive to the
nuisance-model specification used in the doubly robust (DR) evaluator,
we performed a sensitivity analysis in which the logistic-regression
outcome nuisance model was replaced with a HistGradientBoosting classifier.

\subsubsection{Split stability}

To assess sensitivity to the discovery--evaluation split, we repeated the
entire analysis over different independent random splits. Across repetitions,
only the split assignment was varied, and everything else remained unchanged.

\subsubsection{Sensitivity to score-policy encoding}

For individualized teacher and student policies, the intervention budget is
defined among individuals observed in the eligible adverse state ($T=1$). As a sensitivity analysis, we additionally represented the resulting policy
as an explicit score-threshold rule defined over the complete held-out
evaluation cohort. We performed this analysis on the representative discovery--evaluation split while holding everything else fixed.

\subsubsection{Semi-Synthetic Evaluation with the IHDP Dataset}
\label{sec:ihdp_simulation}

The proposed framework is further evaluated using the semi-synthetic Infant Health and Development Program (IHDP) benchmark \cite{louizos2017causal}. Analysis is performed across 10 simulated realizations of the dataset to assess the stability of findings under varying outcome-generating processes. The complete set of variables, their definitions, and the preprocessing procedures are detailed in the supplementary files. The IHDP covariates contain a mixture of continuous and binary or categorical
variables. Accordingly, we replaced K-means with K-Prototypes \cite{kmodes} as the unsupervised subgroup-policy comparator. The number of clusters was selected on the discovery cohort by maximizing the average silhouette coefficient computed from the Gower dissimilarity matrix \cite{gower}. 

In contrast to the PIMA and NHANES analyses, where individual counterfactual outcomes and true treatment effects are unobserved, the semi-synthetic IHDP benchmark provides access to simulated potential outcomes. As a result, ground-truth individual treatment effects are available and can be used to directly evaluate the learned policies. For each simulation, an oracle policy is constructed by prioritizing individuals according to their true treatment benefits under the same intervention-budget constraint as the evaluated policies. The oracle policy thus represents the optimal allocation achievable under the known data-generating mechanism and specified budget.

The policy value achieved by the proposed framework is compared with that of the corresponding oracle policy. This approach enables direct calculation of policy regret, defined as the difference between the oracle policy value and the value achieved by the learned policy. The overlap between individuals selected by the proposed framework and those selected by the oracle policy is also evaluated using policy-allocation agreement measures. Such oracle-based comparisons are not feasible in the PIMA and NHANES applications because the true individual treatment effects and counterfactual outcomes are unknown. The IHDP experiments therefore complement the real-data analyses by providing a controlled setting in which both the value and allocation accuracy of the learned policies can be assessed against known ground truth.

\section{Experimental Setup}\label{setups}
For each observational policy experiment, the policy-specific complete-case
analytic cohort was randomly partitioned into 60\% discovery and 40\%
held-out evaluation data. This discovery–evaluation split follows our previous framework for separating data-adaptive subgroup-policy construction from held-out policy evaluation \cite{acharya2026unsupervised}.

The resulting cohort sizes are
summarized in Table~\ref{tab:policy_cohort_summary}.
\begin{table*}[!h]
\centering
\caption{Analytic cohorts and intervention capacity across observational
policy experiments.}
\scriptsize
\label{tab:policy_cohort_summary}
\begin{tabular}{p{2cm}cccccc}
\toprule
\textbf{Policy experiment}
& \textbf{Analytic $N$}
& \textbf{Discovery $N$}
& \textbf{Evaluation $N$}
& \textbf{Eval\_eligible $N$}
& \textbf{Max\_budget $B_q$} \\
\midrule

PIMA BMI
& 757
& 454
& 303
& 190

& 133 \\

PIMA glucose
& 752
& 451
& 301
& 76

& 53 \\

NHANES smoking history
& 7,217
& 4,330
& 2,887
& 1,273

& 891 \\

\bottomrule
\end{tabular}

\end{table*}

We implemented BCF using the  StochTree \cite{stochtree} library and Causal Forest using EconML \cite{econml}. For the causal-forest teacher, hyperparameters were selected on the discovery
cohort using the out-of-sample $R$-score criterion, since ground-truth
treatment effects are unavailable. Student-model hyperparameters were selected
using five-fold cross-validation to maximize $R^2$ fidelity to the
out-of-fold teacher CATE estimates. In contrast, the BCF teacher was not
data-tuned. We retained the standard literature-guided BCF prior and ensemble
settings following Hahn et al.~\cite{hahn2020bcf}, which regularize the
treatment-effect component separately from the prognostic component
\cite{hahn2020bcf}. Similar BCF-based specifications have also been used in
subsequent applications and extensions \cite{prevot2025hierarchical}. 
The same decision-tree and
random-forest configurations were subsequently retained for the BCF teacher rather than retuned separately, so that changes in
student fidelity and policy performance primarily reflect the change in
teacher signal rather than changes in student-model capacity.

The hyperparameters for the glucose, BMI, smoking history policies and IHDP experiments are tabulated in the table \ref{best_hyperparameters_glucose}, \ref{best_bmi_policy}, \ref{tab:smoking_model_hyperparameters} and \ref{tab:ihdp_model_hyperparameters}.

\begin{table*}[!h]

\scriptsize
\caption{\textbf{PIMA BMI policy:} Model hyperparameters used in the
teacher--student and subgroup-policy analyses. The same decision-tree and
random-forest configurations were subsequently retained for the Bayesian
causal forest teacher rather than retuned separately, so that changes in
student fidelity and policy performance primarily reflect the change in
teacher signal rather than changes in student-model capacity.}
\label{best_bmi_policy}

\begin{adjustwidth}{-1.5cm}{-1.5cm}
\centering
\begin{tabular}{lp{11cm}}
\toprule
\textbf{Model} & \textbf{Hyperparameters used} \\
\midrule

Causal Forest teacher &
$n_{\mathrm{estimators}}=300$,
$\mathrm{max\_depth}=\mathrm{None}$,
$\mathrm{min\_samples\_split}=10$,
$\mathrm{min\_samples\_leaf}=5$ \\

Decision Tree student &
$\mathrm{max\_depth}=6$,
$\mathrm{min\_samples\_leaf}=10$,
$\mathrm{max\_leaf\_nodes}=12$ \\

Random Forest student &
$n_{\mathrm{estimators}}=6$,
$\mathrm{max\_depth}=6$,
$\mathrm{min\_samples\_leaf}=4$,
$\mathrm{max\_leaf\_nodes}=\mathrm{None}$ \\

BCF teacher &
Binary probit outcome model;
prognostic forest: $200$ trees,
$\alpha=0.95$, $\beta=2$,
$\mathrm{min\_samples\_leaf}=5$,
$\mathrm{max\_depth}=8$;
treatment-effect forest: $50$ trees,
$\alpha=0.25$, $\beta=3$,
$\mathrm{min\_samples\_leaf}=5$,
$\mathrm{max\_depth}=5$;
$1000$ Markov chain Monte Carlo (MCMC) draws \\

K-means subgroup policy &
Number of clusters $K=3$ \\
Teacher--student training &
OOF teacher-target generation:
$K_{\mathrm{OOF}}=5$ folds \\
Causal Tree subgroup policy &
Causal splitting criterion: $\mathrm{causal\_mse}$;
$\mathrm{max\_leaf\_nodes}=3$;
honest discovery split: $50\%/50\%$ for structure learning and
leaf-effect estimation;
$5$-fold cross-fitted AIPW leaf-effect scores;\\
\bottomrule
\end{tabular}
\end{adjustwidth}
\end{table*}

\begin{table*}[!h]

\caption{\textbf{PIMA Glucose policy:} Model hyperparameters used in the
teacher--student and subgroup-policy analyses. }
\scriptsize
\begin{adjustwidth}{-1.5cm}{-1.5cm}
\centering
\begin{tabular}{lp{10cm}}
\toprule
\textbf{Model} & \textbf{Hyperparameters / Training Settings} \\
\midrule

Causal Forest Teacher &
$n_{\text{estimators}}=400$,
$\mathrm{max\_depth}=\mathrm{None}$,
$\mathrm{min\_samples\_split}=10$,
$\mathrm{min\_samples\_leaf}=5$ \\

BCF Teacher &
Binary probit outcome model;
prognostic forest: $200$ trees,
$\alpha=0.95$, $\beta=2$,
$\mathrm{min\_samples\_leaf}=5$,
$\mathrm{max\_depth}=8$;
treatment-effect forest: $50$ trees,
$\alpha=0.25$, $\beta=3$,
$\mathrm{min\_samples\_leaf}=5$,
$\mathrm{max\_depth}=5$;
$1000$ MCMC draws \\

Decision Tree &
$\mathrm{max\_depth}=4$,
$\mathrm{min\_samples\_leaf}=10$,
$\mathrm{max\_leaf\_nodes}=12$ \\

Random Forest &
$n_{\text{estimators}}=5$,
$\mathrm{max\_depth}=6$,
$\mathrm{min\_samples\_leaf}=3$,
$\mathrm{max\_leaf\_nodes}=\mathrm{None}$ \\

K-means subgroup policy &
$\mathrm{number\_of\_clusters}=4$ \\
Causal Tree subgroup policy &
Causal splitting criterion: $\mathrm{causal\_mse}$;
$\mathrm{max\_leaf\_nodes}=4$;
honest discovery split: $50\%/50\%$ for structure learning and
leaf-effect estimation;
$5$-fold cross-fitted AIPW leaf-effect scores;
 \\
 Teacher--student training &
OOF teacher-target generation:
$K_{\mathrm{OOF}}=5$ folds \\

\bottomrule
\end{tabular}
\label{best_hyperparameters_glucose}
\end{adjustwidth}
\end{table*}

\begin{table*}[!h]
\centering
\scriptsize
\caption{\textbf{NHANES smoking-history policy:} Model hyperparameters used
in the teacher--student and subgroup-policy analyses.}
\label{tab:smoking_model_hyperparameters}

\begin{adjustwidth}{-1.5cm}{-1.5cm}
\centering
\begin{tabular}{lp{11cm}}
\toprule
\textbf{Model} & \textbf{Hyperparameters used} \\
\midrule

Causal Forest teacher &
$n_{\mathrm{estimators}}=200$,
$\mathrm{max\_depth}=10$,
$\mathrm{min\_samples\_split}=20$,
$\mathrm{min\_samples\_leaf}=20$,
$\mathrm{max\_samples}=0.45$ \\

Decision Tree student &
$\mathrm{max\_depth}=4$,
$\mathrm{min\_samples\_leaf}=25$,
$\mathrm{max\_leaf\_nodes}=10$ \\

Random Forest student &
$n_{\mathrm{estimators}}=8$,
$\mathrm{max\_depth}=5$,
$\mathrm{min\_samples\_leaf}=50$,
$\mathrm{max\_leaf\_nodes}=12$ \\

BCF teacher &
Binary probit outcome model;
prognostic forest: $200$ trees,
$\alpha=0.95$, $\beta=2$,
$\mathrm{min\_samples\_leaf}=5$,
$\mathrm{max\_depth}=8$;
treatment-effect forest: $50$ trees,
$\alpha=0.25$, $\beta=3$,
$\mathrm{min\_samples\_leaf}=5$,
$\mathrm{max\_depth}=5$;
$10$ GFR initialization sweeps and $1000$ MCMC draws \\

K-means subgroup policy &
Number of clusters $K=6$;
$n_{\mathrm{init}}=50$;
K-means++ initialization;
clustering performed on the $7$-component discovery-fitted MCA representation \\

Teacher--student training &
OOF teacher-target generation:
$K_{\mathrm{OOF}}=5$ folds \\ \\

Causal Tree subgroup policy &
Causal splitting criterion: $\mathrm{causal\_mse}$;
$\mathrm{max\_leaf\_nodes}=6$;
honest discovery split: $50\%/50\%$ for structure learning and
leaf-effect estimation;
$5$-fold cross-fitted AIPW leaf-effect scores;
 \\

\bottomrule
\end{tabular}
\end{adjustwidth}
\end{table*}

\begin{table*}[!h]
\centering
\scriptsize
\caption{\textbf{IHDP experiment:} Model hyperparameters and training settings
used in the teacher--student and subgroup-policy analyses.}
\label{tab:ihdp_model_hyperparameters}
\begin{adjustwidth}{-1.5cm}{-1.5cm}
\centering
\begin{tabular}{lp{11cm}}
\toprule
\textbf{Model} & \textbf{Hyperparameters used} \\
\midrule

Causal Forest teacher &
$n_{\mathrm{estimators}}=400$,
$\mathrm{max\_depth}=\mathrm{None}$,
$\mathrm{min\_samples\_leaf}=5$,
$\mathrm{max\_samples}=0.45$,
$\mathrm{max\_features}=\mathrm{None}$ \\

Decision Tree student &
$\mathrm{max\_depth}=3$,
$\mathrm{min\_samples\_leaf}=30$,
$\mathrm{max\_leaf\_nodes}=8$ \\

Random Forest student &
$n_{\mathrm{estimators}}=6$,
$\mathrm{max\_depth}=6$,
$\mathrm{min\_samples\_leaf}=4$ \\

BCF teacher &
Continuous outcome model;
prognostic forest: $200$ trees,
$\alpha=0.95$, $\beta=2$,
$\mathrm{min\_samples\_leaf}=5$,
$\mathrm{max\_depth}=8$;
treatment-effect forest: $50$ trees,
$\alpha=0.25$, $\beta=3$,
$\mathrm{min\_samples\_leaf}=5$,
$\mathrm{max\_depth}=5$;
adaptive coding enabled;
$10$ GFR warm-start draws and $1000$ MCMC draws \\

K-Prototypes subgroup policy &
Number of clusters $K=3$

\\
Teacher--student training &
OOF teacher-target generation:
$K_{\mathrm{OOF}}=5$ folds \\

Causal Tree subgroup policy &
$\mathrm{max\_leaf\_nodes}=3$ (matched to K-Prototypes);
honest discovery split: $50\%/50\%$;
$5$-fold cross-fitted AIPW leaf-effect estimation\\

\bottomrule
\end{tabular}
\end{adjustwidth}
\end{table*}

\section{Results} \label{Results}
We evaluated all PIMA and NHANES dataset-based policies at a 70\% inclusion rate. Following prior work, we view the inclusion rate as a context-specific budget parameter that should be chosen based on external considerations such as available resources, intervention cost, and fairness constraints, rather than as a fixed optimal value determined purely by the model \cite{inclusion_rate1}.

\subsubsection{Prespecified covariate roles across datasets}
\label{subsubsec:covariate_roles}

Table~\ref{tab:covariate_roles} summarizes the prespecified roles of the
pretreatment covariates used in each policy experiment.

\begin{table}[!h]
\centering
\scriptsize

\caption{Prespecified covariate roles used in the policy experiments. For
PIMA and NHANES, these roles were informed by the fixed causal structures
developed in our previous work \cite{acharya2025understanding}. For IHDP,
we did not estimate a causal graph using the proposed causal-discovery
procedure. All available pretreatment covariates were used for
confounding adjustment and treatment-effect modeling. The overlap between
the adjustment-variable and effect-modifier sets is intentional, because a pretreatment covariate may contribute to confounding control while also
modifying the conditional treatment effect.}

\label{tab:covariate_roles}
\begin{tabular}{p{2.1cm}p{2.4cm}p{3.8cm}p{4.0cm}}
\toprule
Dataset/policy
& Treatment
& Adjustment variables
& Candidate effect modifiers \\
\midrule
PIMA--BMI
& Obesity status
& Age, Pregnancies, DiabetesPedigreeFunction(DPF)
& Age, Pregnancies, DPF \\

PIMA--Glucose
& Elevated glucose
& Age, BMI, DPF
& Age, BMI, DPF \\

NHANES--Smoking
& Smoking history
& Gender, Income, Age, Race
&  Gender, Income, Age, Race, Epilepsy, Education\\
\bottomrule
\end{tabular}

\end{table}


\subsection{PIMA Indian Diabetes Dataset-BMI Policy}

\begin{figure*}[!h]
    \centering
\includegraphics[width=10cm,height=5cm]{ 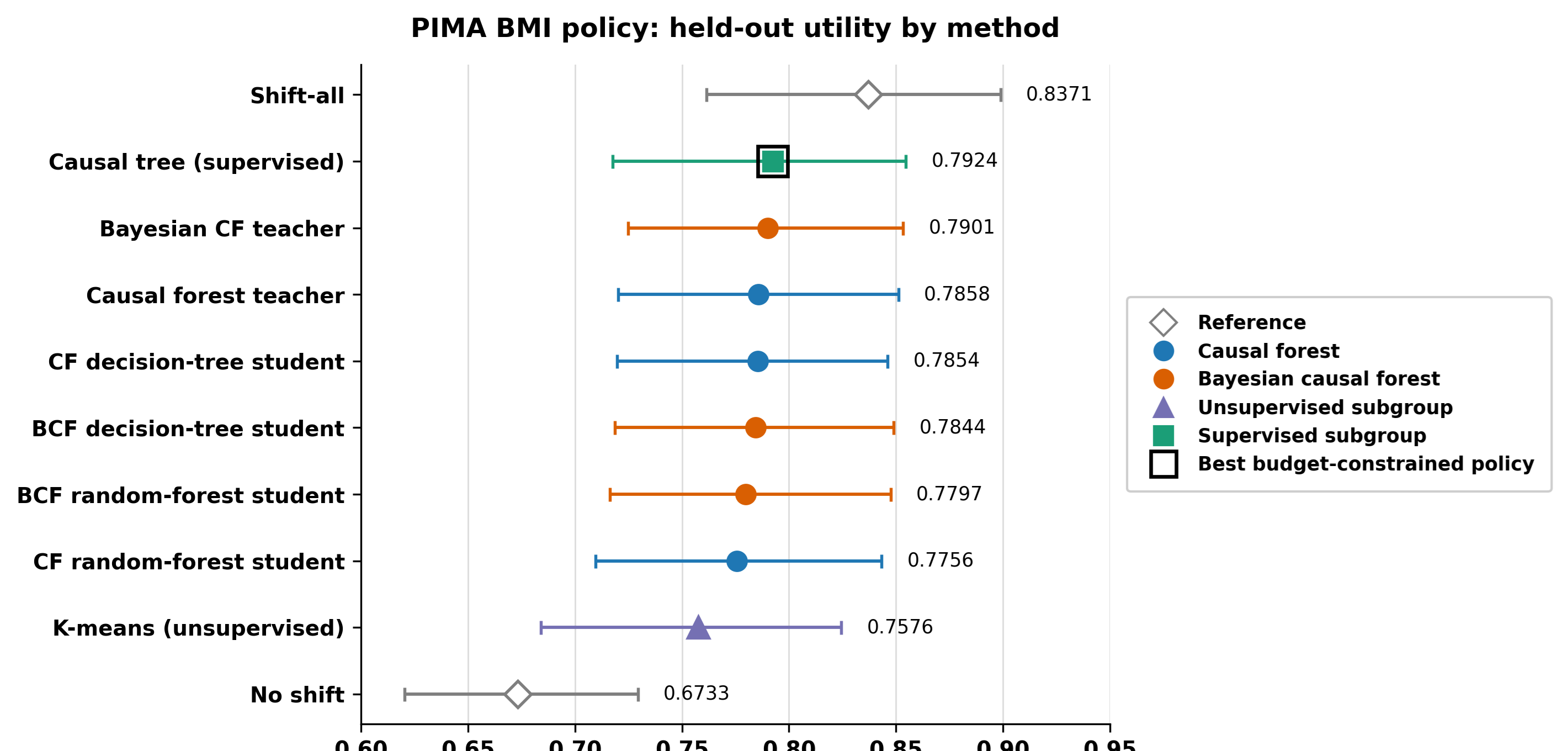}
    \caption{PIMA BMI policy: The causal-tree comparator achieved the highest utility among budget-constrained policies, closely followed by the BCF teacher, causal forest teacher, and distilled decision-tree students. All learned policies outperformed the no-shift reference, while shift-all achieved the highest overall utility}.
    \label{Policy_utility}
\end{figure*}

\begin{figure*}[!h]
    \centering
\includegraphics[width=9cm,height=5cm]{ 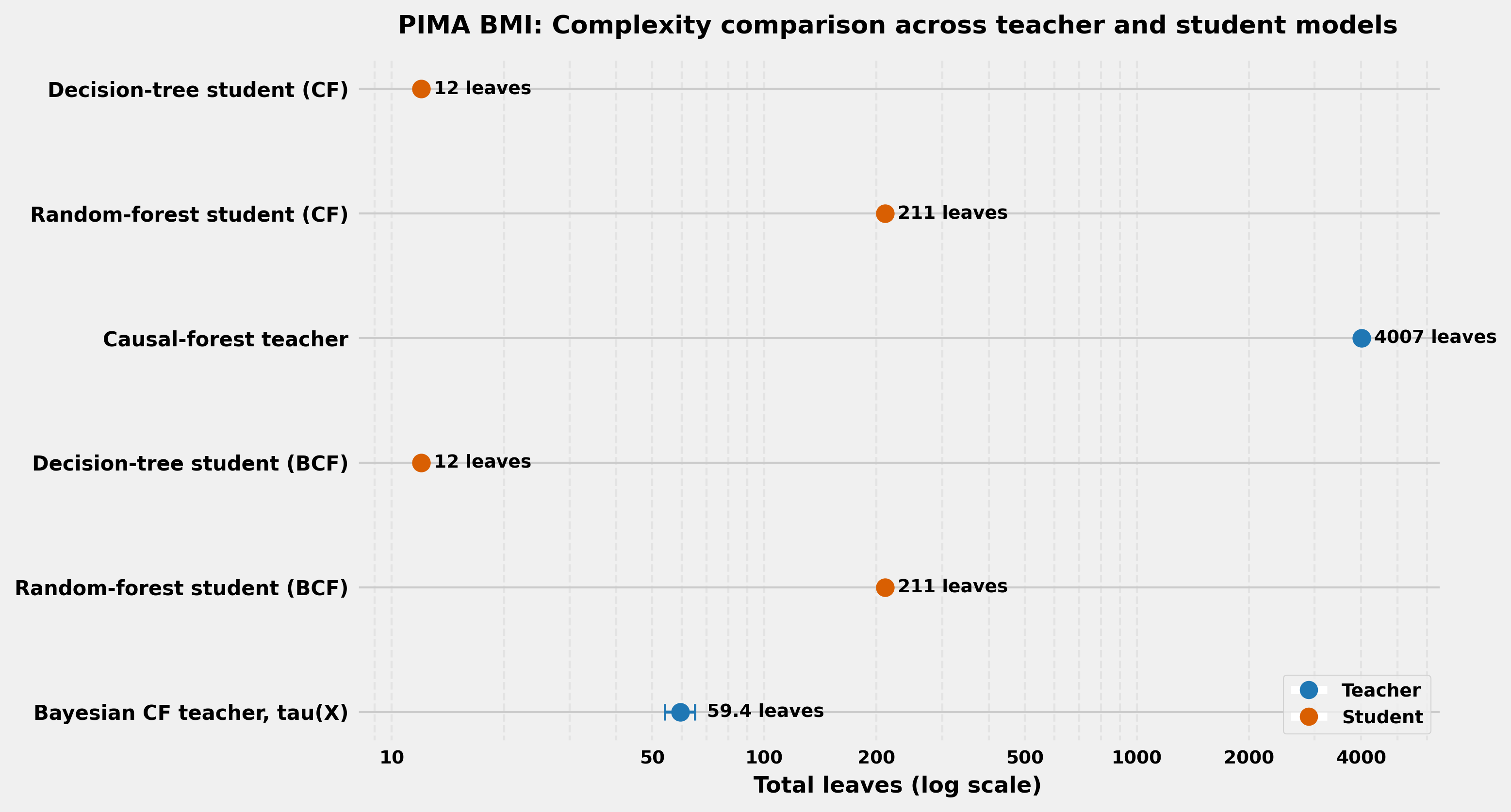}
   \caption{ PIMA BMI policy: Structural complexity of teacher and distilled student models. The distilled decision-tree students provide the most compact representations, while the causal-forest teacher is substantially more complex}.
    \label{complexity}
\end{figure*}

\begin{figure*}[!h]
    \centering
\includegraphics[width=7cm,height=5cm]{ 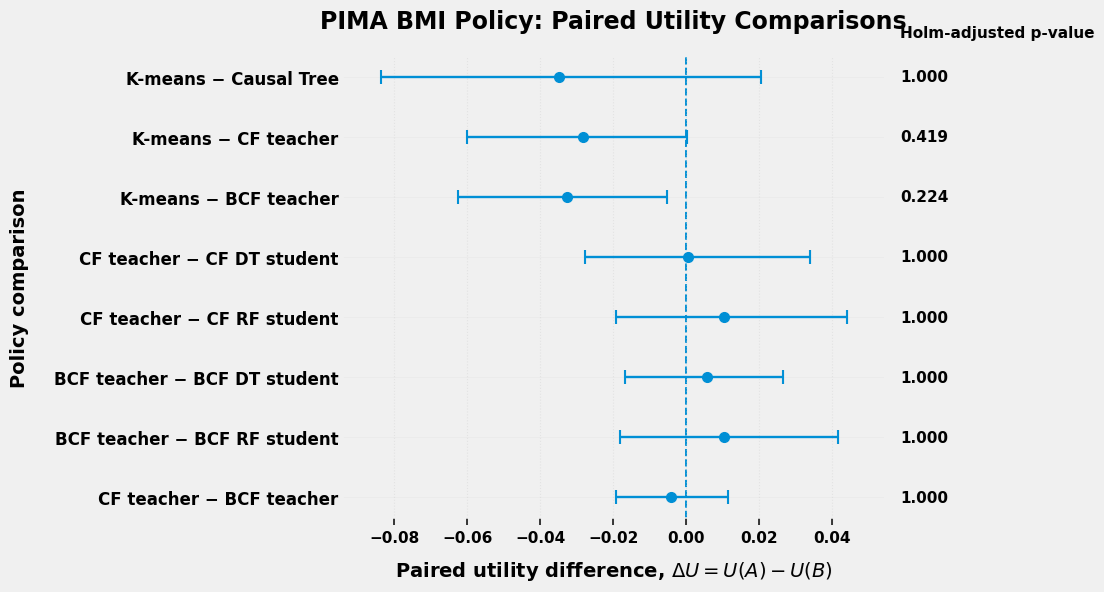}
 \caption{\textbf{PIMA BMI policy: Paired differences in held-out policy
utility at the prespecified 70\% eligible-individual intervention budget.}
}
\label{fig:bmi_paired_utility}
\end{figure*}

We evaluated $K=2$--$9$ on the discovery cohort and selected $K=3$. The complete silhouette-score
profiles and descriptive summaries of the resulting K-means clusters are
provided in the supplementary files. Held-out policy performance, teacher--student fidelity, structural complexity, and paired utility comparisons are summarized in Table~\ref{tab:pima_bmi_teacher_student_policy_performance}, Table~\ref{tab:pima_bmi_teacher_student_fidelity}, and Figs.~\ref{Policy_utility}, \ref{complexity}, and~\ref{fig:bmi_paired_utility}. Teacher--student utility differences were small, and none of the eight prespecified comparisons remained significant after Holm adjustment.

The learned rules as shown in the table \ref{tab:bmi_clinician_student_examples} also admit a plausible clinical interpretation. In the
BMI experiment, several highly prioritized DT and RF regions combined
obesity with higher values of the DPF, a marker of
familial susceptibility to diabetes. This is qualitatively consistent with
the established importance of obesity and family history as diabetes risk
factors \cite{interact2013link,wikner2013risk}. Pregnancy count and age further partitioned these profiles, potentially capturing differences in reproductive and cumulative metabolic
history. This was consistent with large-scale observational evidence that the association between obesity and subsequent diabetes development differs across age groups
\cite{yoo2023differential}.

We emphasize, however, that the numerical thresholds were learned
from the data and should not be interpreted as established clinical cutoffs
or treatment guidelines. Rather, the rules provide an interpretable
description of the heterogeneity learned by the teacher and retained by the
student models. Table~\ref{tab:bmi_bcf_clinician_student_examples} provides representative
clinician-facing examples for the students distilled from the BCF teacher. Despite differences in the exact data-adaptive split thresholds, the
CF- and BCF-distilled students showed qualitatively consistent high-benefit
profiles. In particular, both RF students identified a fully prioritized
region characterized by relatively few pregnancies and high
DPF values. The DT students similarly placed elevated DPF values among their
higher-benefit prioritized leaves. Conversely, very low
DPF values appeared in positive-benefit regions that
fell outside the 70\% intervention capacity for both teacher families.
Thus, the two teachers produced broadly concordant qualitative
prioritization patterns, although their exact student-tree thresholds were
not identical.

\begin{table*}[!h]
\centering

\caption{\textbf{PIMA BMI policy:} Held-out performances. 
Among the budget-constrained learned policies, the Causal Tree subgroup
policy achieved the highest point-estimate utility, followed
closely by the BCF teacher , causal forest
teacher, and their distilled students. K-means achieved a lower
point-estimate utility. The 95\% confidence intervals overlapped
substantially across the learned policies, so the observed ordering should
be interpreted descriptively rather than as evidence of statistically
significant superiority.}

\label{tab:pima_bmi_teacher_student_policy_performance}
\resizebox{\textwidth}{!}{
\begin{tabular}{lllccc}
\toprule
\textbf{Policy family}
& \textbf{Policy}
& \textbf{Distillation source}
& \textbf{Policy risk}
& \textbf{Policy utility}
& \textbf{95\% CI for utility} \\
\midrule

Reference &
Shift all eligible &
---&
0.1629 &
 0.837 &
[0.7613,0.8988]

\\
Reference &
No shift &
---&
0.3267 &
0.6733 &
[0.6205,0.7294]\\

Causal forest
& Teacher
& Causal forest
& 0.2142
& 0.7858
& [0.7201, 0.8511] \\

Causal forest
& Decision-tree student
& Causal forest
& 0.2146
& 0.7854
& [0.7197, 0.8460] \\

Causal forest
& Random-forest student
& Causal forest
& 0.2244
& 0.7756
& [0.7096, 0.8433] \\

\addlinespace

BCF
& Teacher
& BCF
& 0.2099
& 0.7901
& [0.7249, 0.8533] \\

BCF
& Decision-tree student
& BCF
& 0.2156
& 0.7844
& [0.7188, 0.8490] \\

BCF
& Random-forest student
& BCF
& 0.2203
& 0.7797
& [0.7164, 0.8475] \\

\addlinespace

Unsupervised subgroup policy
& K-means
& Not applicable
& 0.2424
& 0.7575
& [0.6841,0.8244]  \\

Supervised subgroup policy
& Causal tree
& Not applicable
& 0.2076
&  0.7923
& [0.7176,0.8545] \\

\bottomrule
\end{tabular}
}
\end{table*}

\begin{table*}[!h]
\centering
\caption{\textbf{PIMA BMI policy:} Teacher--student decision overlap and ranking fidelity at the prespecified eligible-population inclusion rate $q=0.70$. Global agreement was generally high, with decision Jaccard similarities ranging from $0.7500$ to $0.8472$ and eligible-population Spearman correlations ranging from $0.8315$ to $0.8977$. The BCF-distilled decision tree achieved the highest decision overlap, whereas the BCF-distilled random forest achieved the strongest global rank correlation. In contrast, Kendall $\tau_b$ near the allocation boundary was weak or negative for three of the four students, indicating that local ordering around the treatment cutoff was less stable than the overall ranking.}

\label{tab:pima_bmi_teacher_student_fidelity}
\resizebox{\textwidth}{!}{
\begin{tabular}{llccccc}
\toprule
\textbf{Teacher}
& \textbf{Student}
& \textbf{Fidelity ($R^2$)}
& \textbf{Decision Jaccard}

& \textbf{Spearman}
& \textbf{Kendall $\tau_b$}
& \textbf{Boundary Kendall $\tau_b$} \\
\midrule

Causal forest
& Decision tree &

0.7089
 
& 0.7500

& 0.8315
& 0.6805
& 0.1188 \\

Causal forest
& Random forest
& 0.7855
& 0.7733

& 0.8930
& 0.7245
& $-0.1534$ \\

\addlinespace

BCF
& Decision tree
& 0.7306
& 0.8472

& 0.8650
& 0.7138
& $-0.0259$ \\

BCF
& Random forest
 & 0.801
& 0.8095

& 0.8977
& 0.7340
& $-0.1056$ \\

\bottomrule
\end{tabular}
}
\end{table*}

\begin{table*}[!h]
\centering

\caption{\textbf{PIMA BMI policy: Representative clinician-facing examples
from the distilled decision-tree and random-forest students.}
 The shallow RF surrogate showed good fidelity to the full random-forest
predictions ($R^2=0.892$).  }

\label{tab:bmi_clinician_student_examples}

\resizebox{\textwidth}{!}{
\begin{tabular}{p{1.5cm}p{4.2cm}cccp{2.7cm}p{5.5cm}}
\toprule
\textbf{Student}
&
\textbf{Readable rule}
&
\textbf{Eligible $N$}
&
\textbf{Student benefit}
&
\textbf{CF teacher benefit}
&
\textbf{Policy status}
&
\textbf{Illustrative interpretation}
\\
\midrule

DT
&
Pregnancies $\leq3.5$ and
DPF $>0.738$
&
21
&
44.9 pp
&
45.7 pp
&
Fully prioritized
&
This is an exact DT path. An eligible individual satisfying both thresholds
is assigned directly to this leaf, which predicts high positive benefit and
is fully included within the 70\% intervention capacity.
\\

DT
&
Age $>28.5$ and
$0.190<$ DPF $\leq0.426$
&
45
&
23.0 pp
&
23.3 pp
&
Capacity boundary
&
The exact rule identifies a positive-benefit leaf at which the available
capacity is exhausted. The rule therefore identifies the boundary subgroup,
while final allocation follows the prespecified fractional boundary policy.
\\

DT
&
Age $\leq42.5$ and
DPF $\leq0.190$
&
19
&
19.2 pp
&
17.2 pp
&
Positive benefit, outside capacity
&
The rule predicts positive benefit, but this leaf lies below the current
capacity cutoff. The individual is therefore not prioritized at the
prespecified budget despite positive estimated benefit.
\\

\midrule

RF surrogate
&
Pregnancies $\leq3.5$ and
DPF $>0.745$
&
20
&
44.4 pp
&
45.8 pp
&
Fully prioritized RF region
&
This surrogate region summarizes a portion of the full RF in which all
eligible held-out individuals were selected. It provides a concise,
clinician-readable description of the RF behavior, although the actual
policy decision remains that of the full RF ensemble.
\\

RF surrogate
&
$0.190<$ DPF $\leq0.426$
&
80
&
25.1 pp
&
25.2 pp
&
Partially prioritized RF region
&
The surrogate identifies a broad positive-benefit region, but the full RF
makes different policy decisions within it. Thus, the rule explains the
general covariate pattern associated with prioritization without replacing
the full RF decision.
\\

RF surrogate
&
Age $\leq42.5$ and
DPF $\leq0.190$
&
19
&
18.9 pp
&
17.2 pp
&
Positive region, outside capacity
&
The surrogate identifies a region with positive RF benefit, but the full
RF does not prioritize these profiles at the current 70\% capacity.
The rule therefore provides a readable explanation for a lower-priority
region of the RF policy.
\\

\bottomrule
\end{tabular}
}
\end{table*}

\begin{table*}[!h]
\centering

\caption{\textbf{PIMA BMI policy: Representative clinician-facing examples
from the BCF-distilled decision-tree and random-forest students.}
The shallow RF surrogate achieved an $R^2$ of $0.792$ with respect to the
full BCF-distilled RF predictions and is used only for interpretation, not
for policy allocation.}
\label{tab:bmi_bcf_clinician_student_examples}

\resizebox{\textwidth}{!}{
\begin{tabular}{p{1.5cm}p{4.5cm}cccp{2.7cm}p{5.6cm}}
\toprule

\textbf{Student}
&
\textbf{Readable rule}
&
\textbf{Eligible $N$}
&
\textbf{Student benefit}
&
\textbf{BCF teacher benefit}
&
\textbf{Policy status}
&
\textbf{Illustrative interpretation}
\\
\midrule

DT
&
$0.5<$ Pregnancies $\leq5.5$ and
$0.589<$ DPF $\leq1.280$
&
23
&
33.1 pp
&
34.3 pp
&
Fully prioritized
&
This is an exact DT path. An eligible individual satisfying these thresholds
is assigned directly to this terminal leaf, which predicts positive benefit
and is fully included within the prespecified 70\% intervention capacity.
\\

DT
&
Age $>40.5$, Pregnancies $>5.5$, and
DPF $>0.304$
&
15
&
22.5 pp
&
22.8 pp
&
Capacity boundary
&
The exact DT rule identifies a positive-benefit leaf at which the intervention
capacity is exhausted. At the prespecified budget, 60\% of eligible
individuals in this leaf receive allocation through the boundary mechanism.
\\

DT
&
Age $>24.5$ and
DPF $\leq0.182$
&
21
&
19.8 pp
&
19.0 pp
&
Positive benefit, outside capacity
&
The rule predicts positive benefit, but the leaf falls below the current
capacity cutoff. 
\\

\midrule

RF surrogate
&
Pregnancies $\leq4.5$ and
DPF $>0.735$
&
24
&
36.3 pp
&
36.9 pp
&
Fully prioritized RF region
&
The surrogate identifies a readable region in which all eligible evaluation
profiles were selected by the full RF. 
\\

RF surrogate
&
Age $\leq40.5$ and
$0.318<$ DPF $\leq0.735$
&
67
&
28.7 pp
&
29.5 pp
&
Partially prioritized RF region
&
The surrogate identifies a positive-benefit region in which the full RF
prioritized approximately 98.5\% of eligible individuals. .
\\

RF surrogate
&
Age $>24.5$ and
DPF $\leq0.182$
&
21
&
20.4 pp
&
19.0 pp
&
Positive region, outside capacity
&
The surrogate identifies a region with positive mean RF benefit, but none
of the eligible profiles in this region are prioritized under the current
70\% intervention capacity.
\\

\bottomrule
\end{tabular}
}
\end{table*}


\subsection{PIMA Indian Diabetes Dataset-Glucose Policy}

\begin{figure*}[!h]
    \centering
\includegraphics[width=10cm,height=5cm]{ 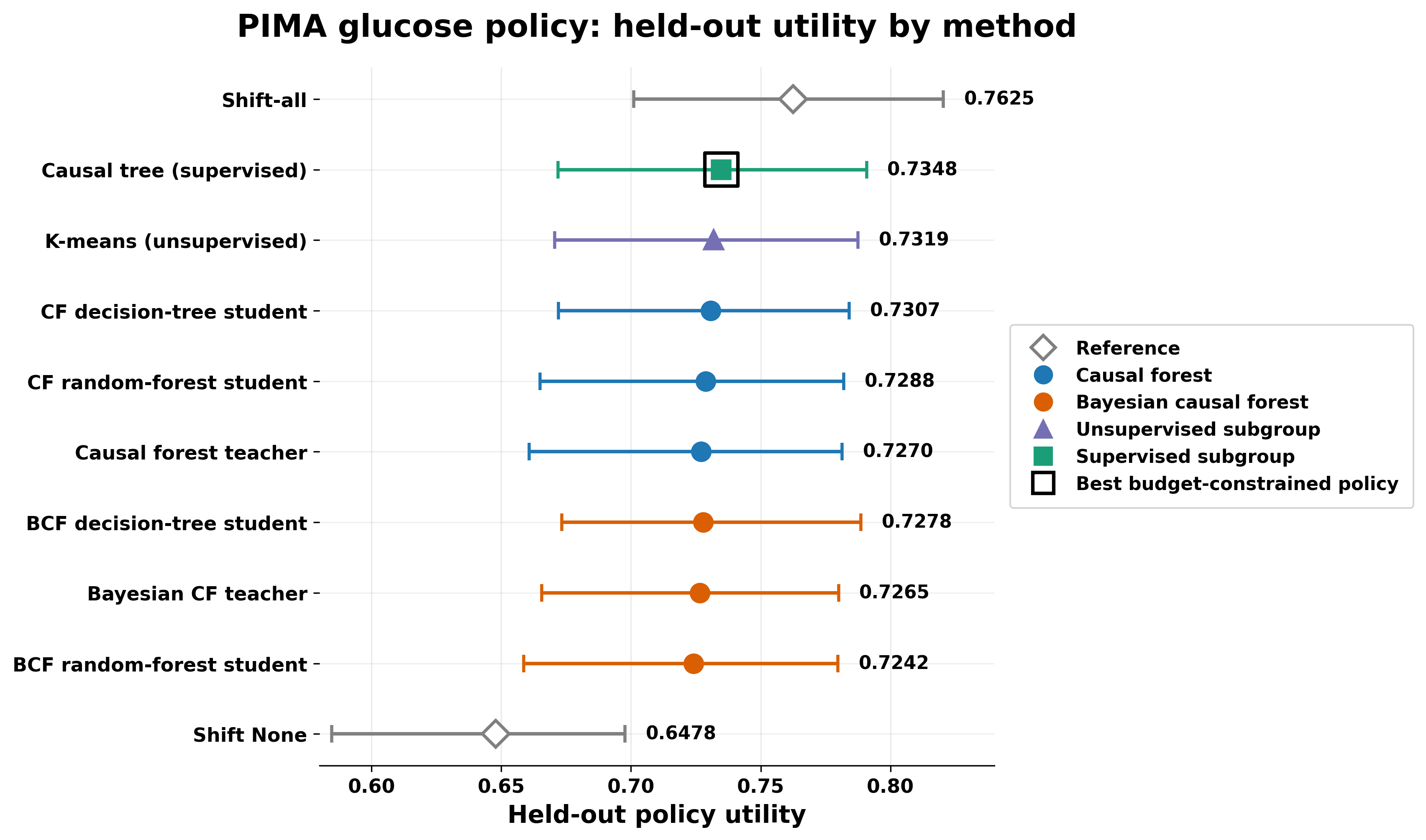}
  
\caption{\textbf{PIMA glucose policy:} The causal-tree comparator achieved the
highest utility among the budget-constrained learned policies, followed closely
by K-means and the causal-forest-distilled decision-tree student. }
    \label{Policy_utility_glucose}
\end{figure*}

\begin{figure*}[!h]
    \centering
\includegraphics[width=9cm,height=5cm]{ 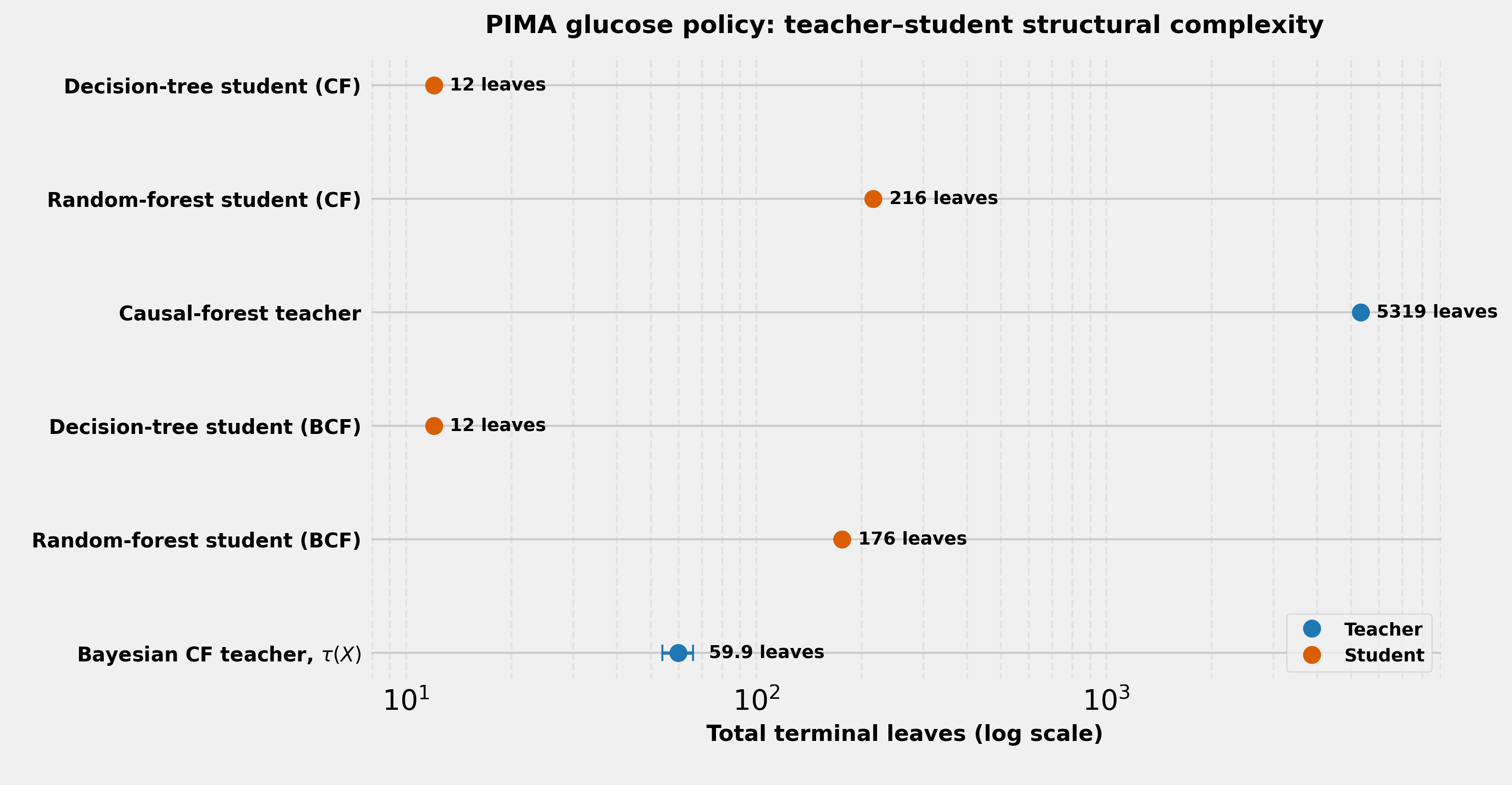}
\caption{\textbf{PIMA glucose policy: Teacher--student structural complexity.} Both distilled student models exhibit substantially fewer terminal leaves compared to the causal-forest teacher. The decision-tree student is also more compact than the BCF treatment-effect forest, according to the reported per-draw measure, while the BCF-distilled random forest has more leaves.}
\label{fig:glucose_complexity}
\end{figure*}

\begin{figure*}[!h]
    \centering
\includegraphics[width=7cm,height=5cm]{ 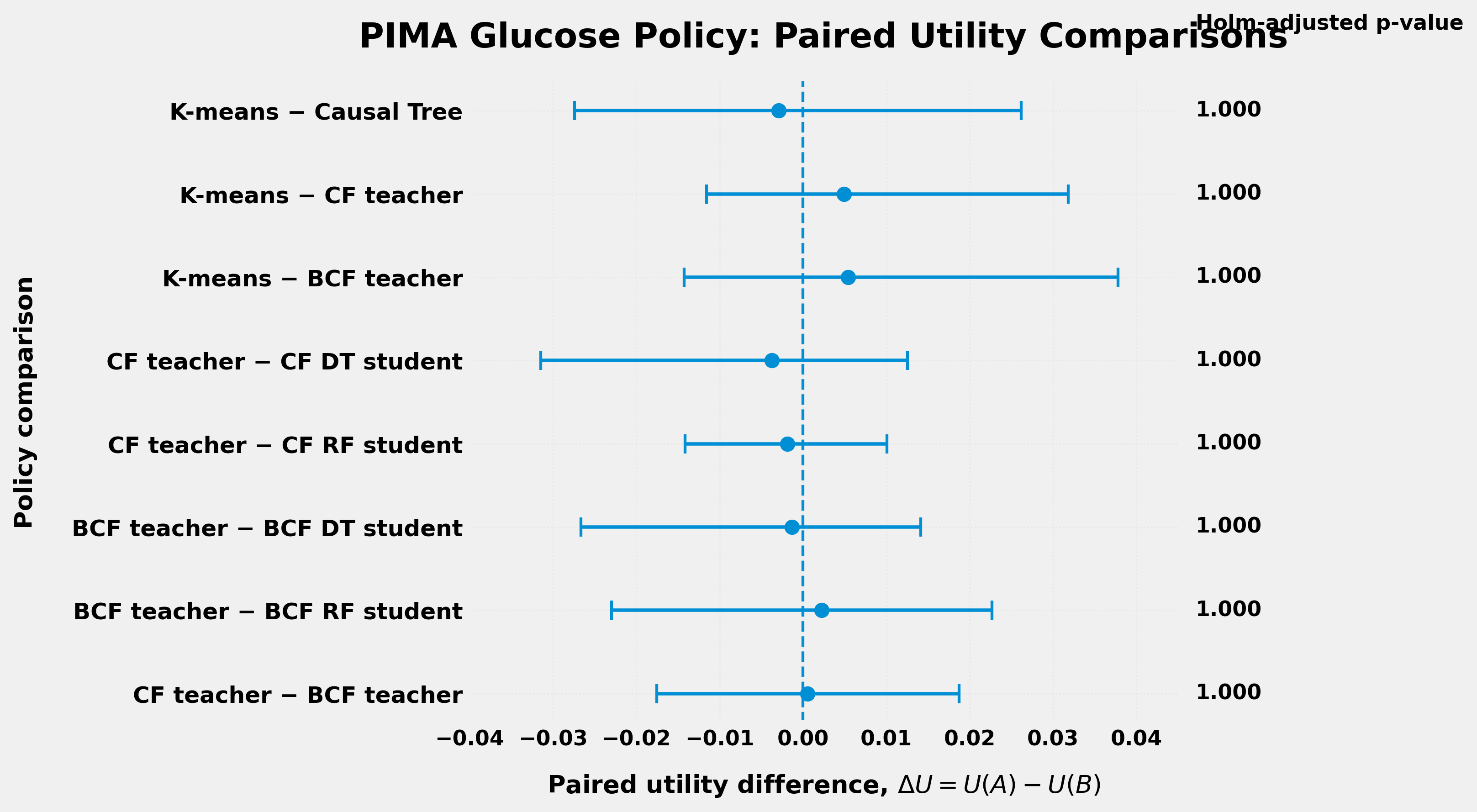}

\caption{\textbf{PIMA glucose policy: Paired differences in held-out policy
utility at the prespecified eligible-individual intervention budget.}
Positive values favor policy $A$ and
negative values favor policy $B$. Holm-adjusted $p$-values are shown at the
right to account for multiplicity across the eight prespecified comparisons.
All confidence intervals include zero, and none of the comparisons are
statistically significant after Holm adjustment.}

\label{fig:glucose_paired_utility}
\end{figure*}

Table~\ref{tab:pima_glucose_teacher_student_policy_performance} and Fig.~\ref{Policy_utility_glucose} summarize held-out glucose-policy performance. Utilities were similar across the learned policies, with substantially overlapping 95\% confidence intervals. Teacher--student decision overlap was generally high despite the simpler student models (Table~\ref{tab:pima_glucose_teacher_student_fidelity}), although prediction fidelity, ranking agreement, and policy overlap were not necessarily maximized by the same student. Distillation substantially reduced structural complexity, and none of the prespecified paired utility comparisons remained significant after Holm adjustment (Figs.~\ref{fig:glucose_complexity} and~\ref{fig:glucose_paired_utility}). The learned rules for the glucose policy
(Tables~\ref{tab:glucose_bcf_clinician_student_examples}
and~\ref{tab:glucose_cf_clinician_student_examples})
were broadly compatible with established diabetes physiology. Across both the
causal-forest and Bayesian-causal-forest distillation analyses, higher-priority
regions tended to involve moderately high to high BMI values, consistent with
the well-established relationship between adiposity, insulin resistance, and
type 2 diabetes risk \cite{kahn2006mechanisms}. This BMI-related pattern was
therefore relatively stable to the choice of teacher. Under the Bayesian
causal forest, however, the association was not strictly monotonic. The
highest-BMI decision-tree leaf (BMI $>43.100$ and DPF $\leq1.011$) had a lower
mean estimated benefit (28.8 pp) than the adjacent moderately high-BMI leaf
(BMI $33.250$--$43.100$; 32.4 pp). Because the former leaf contained only six
eligible individuals and the corresponding random-forest surrogate did not
reproduce this decrease, we interpret it cautiously as possible local
heterogeneity or estimation variability rather than evidence of a biological
threshold.

DPF also appeared repeatedly in the learned rules, indicating that familial
susceptibility contributed to the estimated heterogeneity in benefit, but its
role was less stable across teachers. Under the BCF, DPF
acted primarily as a broad threshold condition, with the high-DPF region among
those examined showing comparatively lower estimated benefit. Under the
causal forest, DPF appeared more frequently as a splitting variable and its
relationship with estimated benefit depended on combinations with age and BMI
rather than following a single monotonic direction. We therefore regard the
BMI-related pattern as the more reproducible cross-teacher feature, whereas
the specific DPF pattern observed under the BCF is better
interpreted as teacher-dependent and hypothesis-generating rather than as an
established physiological finding. It is important to note that the numerical
thresholds in these rules are data-driven partitions of the estimated
treatment-effect surface and should not be interpreted as clinical cutoffs.

\begin{table*}[!h]
\centering

\caption{\textbf{PIMA glucose policy:} Held-out performance of the
causal-forest and Bayesian-causal-forest teachers, their distilled
decision-tree and random-forest students, and the K-means and causal-tree
subgroup policies. Among the learned policies, the causal-tree subgroup
policy achieved the highest point-estimate utility, followed
closely by K-means and the causal-forest-distilled decision-tree
student. The remaining teacher and student policies also produced
similar point estimates. The 95\% confidence intervals overlapped across all learned policies.}

\label{tab:pima_glucose_teacher_student_policy_performance}

\resizebox{\textwidth}{!}{
\begin{tabular}{lllccc}
\toprule
\textbf{Policy family}
& \textbf{Policy}
& \textbf{Distillation source}
& \textbf{Policy risk}
& \textbf{Policy utility}
& \textbf{95\% CI for utility} \\
\midrule

Reference
& Shift all eligible
& ---
& 0.2375
& 0.7625
& [0.7010, 0.8204] \\

Reference
& No shift
& ---
& 0.3522
& 0.6478
& [0.5847, 0.6977] \\

\addlinespace

Causal forest
& Teacher
& Causal forest
& 0.2730
& 0.7270
& [0.6607, 0.7814] \\

Causal forest
& Decision-tree student
& Causal forest
& 0.2693
& 0.7307
& [0.6720, 0.7841] \\

Causal forest
& Random-forest student
& Causal forest
& 0.2712
& 0.7288
& [0.6649, 0.7820] \\

\addlinespace

BCF
& Teacher
& BCF
& 0.2735
& 0.7265
& [0.6656, 0.7800] \\

BCF
& Decision-tree student
& BCF
& 0.2722
& 0.7278
& [0.6733, 0.7886] \\

BCF
& Random-forest student
& BCF
& 0.2758
& 0.7242
& [0.6586, 0.7797] \\

\addlinespace

Unsupervised subgroup policy
& K-means
& Not applicable
& 0.2681
& 0.7319
& [0.6706, 0.7875] \\

Supervised subgroup policy
& Causal tree
& Not applicable
& 0.2652
& 0.7348
& [0.6719, 0.7908] \\

\bottomrule
\end{tabular}
}
\end{table*}

\begin{table*}[!h]
\centering
\caption{\textbf{PIMA glucose policy:} Teacher--student decision overlap and
ranking fidelity at the prespecified eligible-population inclusion rate. Decision overlap was generally high, with policy Jaccard similarities
ranging from $0.7377$ to $0.8596$. The CF-distilled random forest and
BCF-distilled decision tree achieved the highest decision overlap, while the CF-distilled random forest showed the strongest global
rank agreement with its teacher. The BCF-distilled random forest achieved the highest
teacher--student prediction fidelity ($R^2=0.7634$), although its policy
overlap and rank correlations were lower than those of the BCF-distilled
decision tree. }

\label{tab:pima_glucose_teacher_student_fidelity}

\resizebox{\textwidth}{!}{
\begin{tabular}{llccccc}
\toprule

\textbf{Teacher}
& \textbf{Student}
& \textbf{Fidelity ($R^2$)}
& \textbf{Decision Jaccard}
& \textbf{Spearman}
& \textbf{Kendall $\tau_b$}
& \textbf{Boundary Kendall $\tau_b$} \\

\midrule

Causal forest
& Decision tree
& 0.6108
& 0.7377
& 0.7313
& 0.5672
& 0.402911 \\

Causal forest
& Random forest
& 0.7078
& 0.8596
& 0.8007
& 0.6118
&0.642857 \\  

\addlinespace

BCF
& Decision tree
& 0.7398
& 0.8596
& 0.7234
& 0.5889
&  0.267261 \\

BCF
& Random forest
& 0.7634
& 0.7667
& 0.6778
& 0.5435
& -0.254588 \\

\bottomrule
\end{tabular}
}

\vspace{1mm}

\end{table*}

\begin{table*}[!h]
\centering

\caption{\textbf{PIMA glucose policy: Representative clinician-facing examples
from the causal-forest-distilled decision-tree and random-forest students.}
The shallow RF surrogate achieved an $R^2$ of $0.782$ with respect to the
full causal-forest-distilled RF predictions and is used only for
interpretation, not for policy allocation.}
\label{tab:glucose_cf_clinician_student_examples}

\resizebox{\textwidth}{!}{
\begin{tabular}{p{1.5cm}p{4.8cm}cccp{3.0cm}p{5.8cm}}
\toprule

\textbf{Student}
&
\textbf{Readable rule}
&
\textbf{Eligible $N$}
&
\textbf{Student benefit}
&
\textbf{CF teacher benefit}
&
\textbf{Policy status}
&
\textbf{Illustrative interpretation}
\\
\midrule

DT
&
$33.700 < \mathrm{BMI} \leq 41.800$ and
DPF $\leq 0.440$
&
13
&
37.1 pp
&
38.8 pp
&
Fully prioritized
&
This is an exact DT path. An eligible individual satisfying these thresholds
is assigned directly to this terminal leaf, which predicts positive benefit
and is fully included within the prespecified 70\% intervention capacity.
\\

DT
&
Age $\leq48.000$, BMI $>30.750$, and
DPF $>0.440$
&
21
&
27.5 pp
&
28.7 pp
&
Capacity boundary
&
This exact DT leaf lies at the intervention-capacity boundary. Individuals
within the leaf receive the same DT benefit score, and approximately 42.9\%
of eligible profiles in this leaf are prioritized through the boundary
allocation mechanism.
\\

DT
&
$25.500 < \mathrm{Age} \leq 46.500$ and
$26.750 < \mathrm{BMI} \leq 30.750$
&
9
&
24.0 pp
&
20.8 pp
&
Positive benefit, outside capacity
&
The exact DT rule predicts positive benefit, but the leaf falls below the
current capacity cutoff and none of its eligible evaluation profiles are
prioritized at the prespecified budget.
\\

\midrule

RF surrogate
&
$33.750 < \mathrm{BMI} \leq 43.050$ and
DPF $\leq 0.440$
&
12
&
37.0 pp
&
39.1 pp
&
Fully prioritized RF region
&
The surrogate identifies a readable region in which all eligible evaluation
profiles were selected by the full RF. Because this is a surrogate region,
the rule summarizes rather than exactly reproduces the ensemble.
\\

RF surrogate
&
$24.500 < \mathrm{Age} \leq 46.500$ and
BMI $\leq33.750$
&
19
&
28.0 pp
&
27.1 pp
&
Partially prioritized RF region
&
The surrogate identifies a positive-benefit region in which the full RF
prioritizes approximately 47.4\% of eligible profiles, demonstrating
within-region variation in individual RF allocation decisions.
\\

RF surrogate
&
Age $\leq24.500$ and
$30.150 < \mathrm{BMI} \leq 33.750$
&
3
&
24.8 pp
&
23.8 pp
&
Positive region, outside capacity
&
The surrogate identifies a region with positive mean RF-estimated benefit,
but none of the eligible profiles in this region are prioritized under the
current 70\% intervention capacity.
\\

\bottomrule
\end{tabular}
}
\end{table*}

\begin{table*}[!h]
\centering

\caption{\textbf{PIMA glucose policy: Representative clinician-facing examples
from the BCF-distilled decision-tree and random-forest students.}
 The shallow RF surrogate achieved an $R^2$ of $0.899$ with respect to the
full BCF-distilled RF predictions and is used only for interpretation, not
for policy allocation.}
\label{tab:glucose_bcf_clinician_student_examples}

\resizebox{\textwidth}{!}{
\begin{tabular}{p{1.5cm}p{4.8cm}cccp{3.0cm}p{5.8cm}}
\toprule

\textbf{Student}
&
\textbf{Readable rule}
&
\textbf{Eligible $N$}
&
\textbf{Student benefit}
&
\textbf{BCF teacher benefit}
&
\textbf{Policy status}
&
\textbf{Illustrative interpretation}
\\
\midrule

DT
&
$33.250 < \mathrm{BMI} \leq 43.100$ and
DPF $\leq 1.011$
&
30
&
32.4 pp
&
32.7 pp
&
Fully prioritized
&
This is an exact DT path. An eligible individual satisfying these thresholds
is assigned directly to this terminal leaf, which predicts positive benefit
and is fully included within the prespecified 70\% intervention capacity.
\\

DT
&
Age $>24.500$,
$28.750 < \mathrm{BMI} \leq 33.250$, and
DPF $\leq 1.011$
&
19
&
30.0 pp
&
31.0 pp
&
Capacity boundary
&
This exact DT leaf lies at the intervention-capacity boundary. Individuals
within the leaf receive the same DT benefit score, and approximately 94.7\%
of eligible profiles in this leaf are prioritized under the boundary
allocation mechanism.
\\

DT
&
BMI $>43.100$ and
DPF $\leq 1.011$
&
6
&
28.8 pp
&
27.6 pp
&
Positive benefit, outside capacity
&
The exact DT rule predicts positive benefit, but the leaf falls below the
current capacity cutoff and none of its eligible evaluation profiles are
prioritized at the prespecified budget.
\\

\midrule

RF surrogate
&
BMI $>32.850$ and
DPF $\leq 1.086$
&
38
&
31.9 pp
&
31.8 pp
&
Predominantly prioritized RF region
&
The surrogate identifies a readable positive-benefit region in which the full
RF prioritizes approximately 94.7\% of eligible profiles. Because this is a
surrogate region, individual RF decisions may still differ within the region.
\\

RF surrogate
&
$28.750 < \mathrm{BMI} \leq 32.850$ and
DPF $\leq 1.086$
&
19
&
29.7 pp
&
31.0 pp
&
Partially prioritized RF region
&
The surrogate identifies a positive-benefit region in which the full RF
prioritizes approximately 68.4\% of eligible profiles, illustrating
within-region variation in the ensemble's individual allocation decisions.
\\

RF surrogate
&
BMI $>28.750$ and
DPF $>1.086$
&
6
&
21.2 pp
&
18.2 pp
&
Positive region, outside capacity
&
The surrogate identifies a region with positive mean RF-estimated benefit,
but none of the eligible profiles in this region are prioritized under the
current 70\% intervention capacity.
\\

\bottomrule
\end{tabular}
}
\end{table*}

\subsection{NHANES Smoking Policy}
\begin{figure}[!h]
    \centering
 \includegraphics[width=7cm,height=5cm]{ 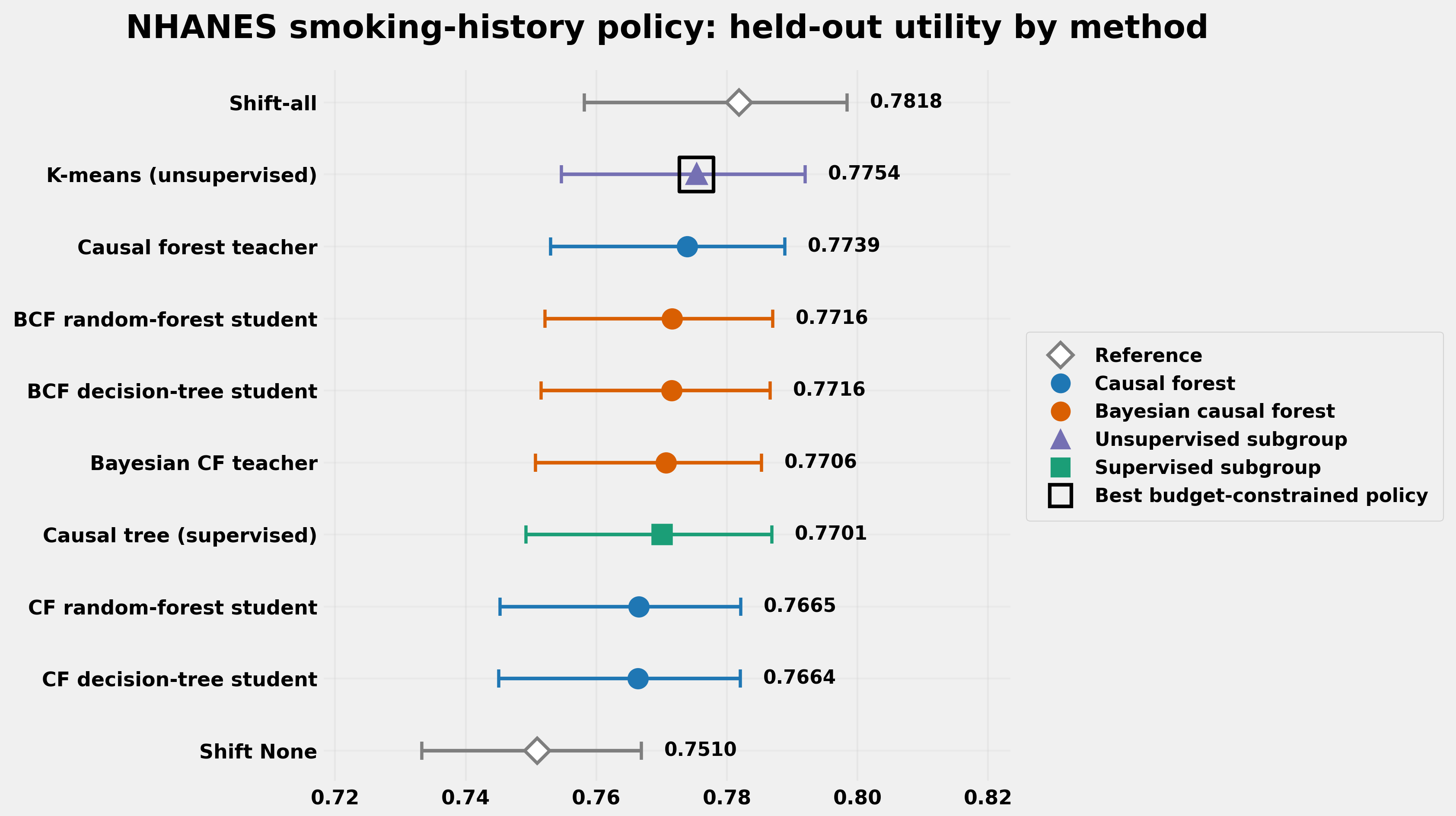}
    
    \caption{\textbf{NHANES smoking-history policy: held-out utility by method.}
    Among the budget-constrained
    learned policies, K-means achieved the highest point-estimate utility
    ($0.7754$), followed by the causal-forest teacher ($0.7739$).
   }
    \label{fig:smoking_policy_utility}
\end{figure}

\begin{figure}[!h]
    \centering
   \includegraphics[width=9cm,height=5cm]{ 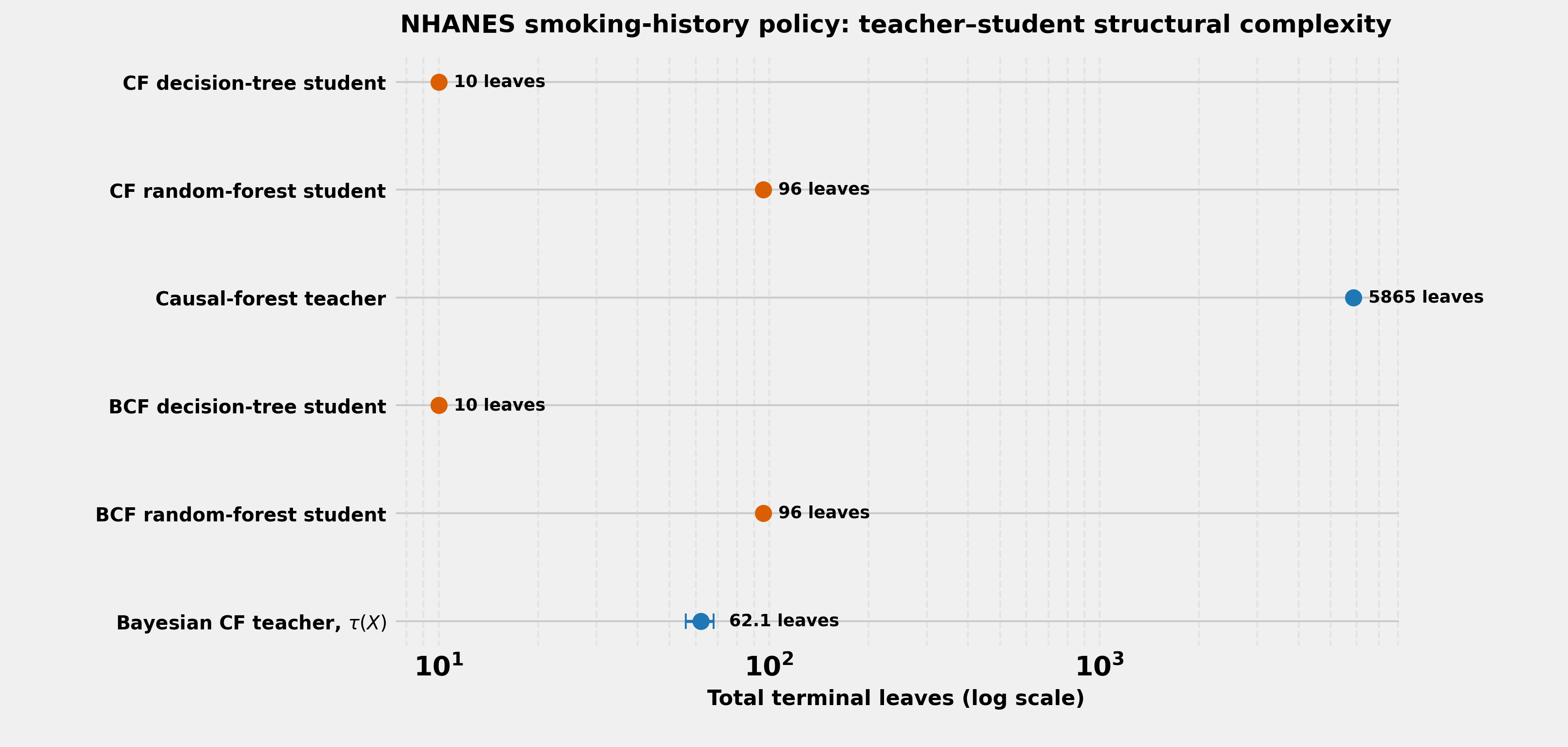}
    \caption{\textbf{NHANES smoking-history policy: teacher--student structural
    complexity.} Structural complexity is summarized using the total number
    of terminal leaves on a logarithmic scale. Variation in realized leaf count of BCF reflects the adaptively learned posterior tree structures.}
    \label{fig:smoking_model_complexity}
\end{figure}

\begin{figure}[!h]
    \centering
  \includegraphics[width=7cm,height=5cm]{ 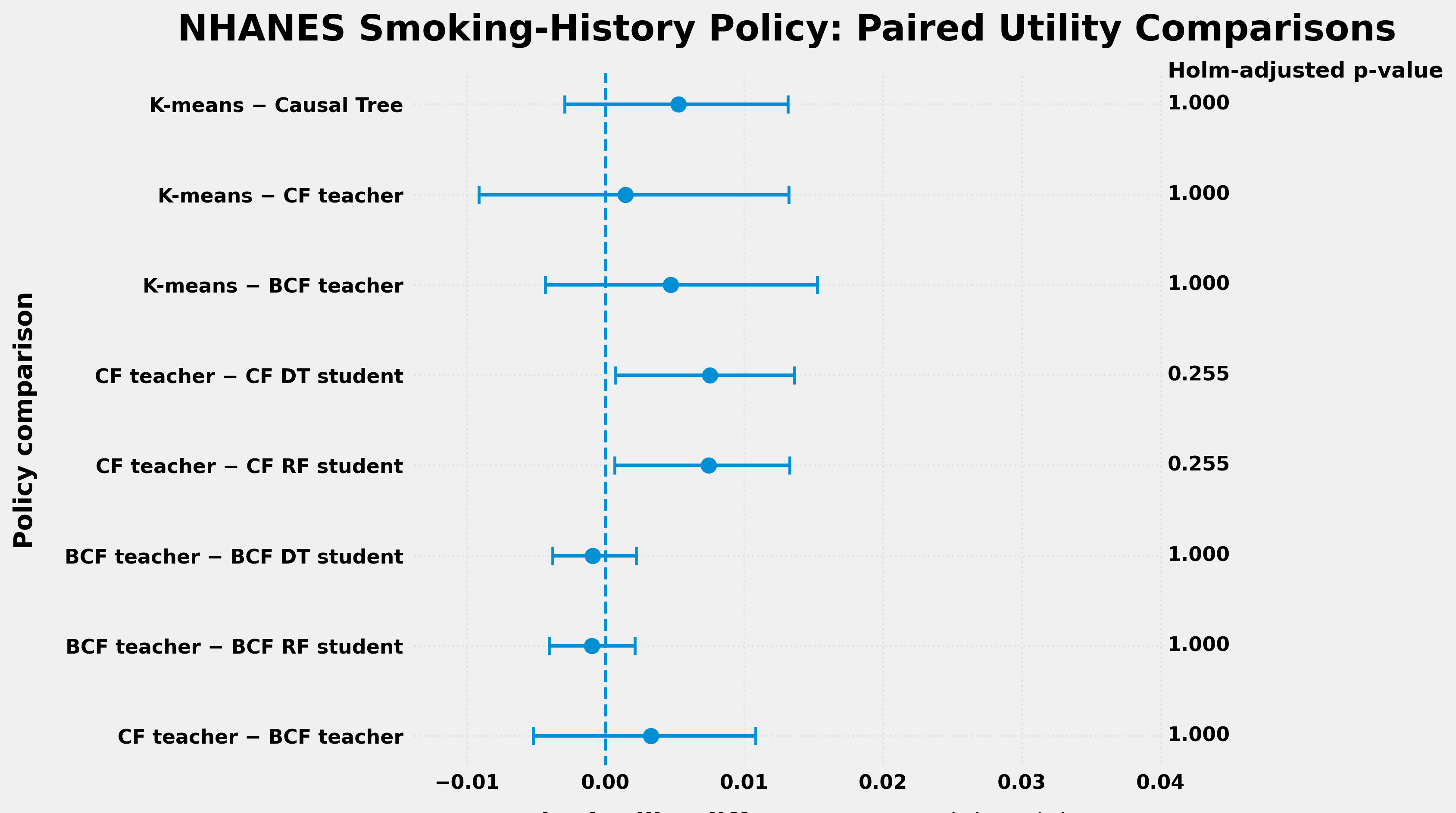}
    \caption{\textbf{NHANES smoking-history policy: paired utility comparisons.}
    Paired held-out utility differences are shown for the prespecified policy
    comparisons, with 95\% paired-bootstrap confidence intervals and
    Holm-adjusted $p$-values. Differences are defined as
    $\Delta U = U(A)-U(B)$, such that positive values favor the first policy
    named in each comparison. None of the comparisons remained
    statistically significant after Holm adjustment at the 0.05 level.}
    \label{fig:smoking_paired_utility}
\end{figure}

Held-out smoking-policy performance, teacher--student fidelity, structural complexity, and paired utility comparisons are summarized in Table~\ref{tab:smoking_teacher_student_policy_performance}, Table~\ref{tab:smoking_teacher_student_fidelity}, and Figs.~\ref{fig:smoking_policy_utility}--\ref{fig:smoking_paired_utility}. Utilities were similar across the learned policies, while the BCF-distilled students showed strong teacher agreement with substantially simpler representations. None of the prespecified paired comparisons remained significant after Holm adjustment.

Across the clinician-facing rules as shown in the table \ref{tab:smoking_cf_clinician_rules} and \ref{tab:smoking_bcf_clinician_rules}, age emerged as a consistent policy-relevant feature across both causal teachers. In the CF-distilled students, fully prioritized regions generally involved individuals aged 60 years or younger, whereas positive-benefit regions among those older than 60 years appeared near or outside the allocation boundary. A similar pattern was evident for the BCF-distilled students, in which predicted benefit decreased across the illustrative age strata from younger than 40 years to 40--60 years and then to older than 60 years. This qualitative agreement suggests that age-related heterogeneity was recovered under both teacher models, although the specific subgroup definitions differed. Race/ethnicity and educational attainment also appeared in the rules derived from both teachers, while income was prominent in the CF-derived rules and gender appeared in the BCF-derived rules. We interpret these demographic and socioeconomic partitions in the context of established social and structural determinants of sleep-health disparities \cite{grandner2016sleep}, rather than as evidence of direct biological mechanisms. We regard these finer subgroup distinctions as model-dependent rather than established clinical findings. More generally, the reported thresholds are data-driven decision boundaries and should not be interpreted as validated clinical cutoffs.

\begin{table*}[!h]
\centering

\caption{\textbf{NHANES smoking-history policy:} Held-out performances. Among the learned
policies evaluated under the $q=0.70$ maximum eligible-population budget,
K-means achieved the highest point-estimate utility ($0.7754$), followed
closely by the causal-forest teacher ($0.7739$). The Bayesian causal-forest
teacher and its distilled students achieved similar utility
($0.7706$--$0.7716$), while the Causal Tree achieved utility $0.7701$.
The shift-all policy is shown only as an unconstrained reference and shifts
all eligible individuals. Confidence intervals substantially overlapped
across the learned policies.}
\label{tab:smoking_teacher_student_policy_performance}

\resizebox{\textwidth}{!}{
\begin{tabular}{lllccc}
\toprule
\textbf{Policy family}
& \textbf{Policy}
& \textbf{Distillation source}
& \textbf{Policy risk}
& \textbf{Policy utility}
& \textbf{95\% CI for utility} \\
\midrule

Reference
& Shift all eligible
& ---
& 0.2182
& 0.7818
& [0.7581, 0.7984] \\

Reference
& No shift
& ---
& 0.2490
& 0.7510
& [0.7333, 0.7669] \\

\addlinespace

Causal forest
& Teacher
& Causal forest
& 0.2261
& 0.7739
& [0.7530, 0.7889] \\

Causal forest
& Decision-tree student
& Causal forest
& 0.2336
& 0.7664
& [0.7450, 0.7820] \\

Causal forest
& Random-forest student
& Causal forest	
& 0.2335
& 0.7665
& [0.7453, 0.7821] \\

\addlinespace

BCF
& Teacher
& BCF
& 0.2294
& 0.7706
& [0.7507, 0.7853] \\

BCF
& Decision-tree student
& BCF
& 0.2284
& 0.7716
& [0.7515, 0.7866] \\

BCF
& Random-forest student
& BCF
& 0.2284
& 0.7716
& [0.7522, 0.7870] \\

\addlinespace

Unsupervised subgroup policy
& K-means
& Not applicable
& 0.2246
& 0.7754
& [0.7546, 0.7920] \\

Supervised subgroup policy
& Causal Tree
& Not applicable
& 0.2299
& 0.7701
& [0.7492, 0.7869] \\

\bottomrule
\end{tabular}
}
\end{table*}

\begin{table*}[!h]
\centering

\caption{\textbf{NHANES smoking-history policy:} Teacher--student decision
overlap and ranking fidelity at the prespecified eligible-population inclusion
rate $q=0.70$. Teacher--student decision overlap was high across all distilled
policies, with Jaccard similarities ranging from $0.7749$ to $0.9370$.
The BCF-distilled random forest achieved the highest decision overlap, the strongest global rank agreement with its teacher, and the highest
boundary-region rank agreement. The BCF-distilled decision
tree also showed strong agreement with its teacher. Boundary-region rank agreement was positive for the
BCF-distilled students but negative for the CF-distilled students.}

\label{tab:smoking_teacher_student_fidelity}

\resizebox{\textwidth}{!}{
\begin{tabular}{llccccc}
\toprule

\textbf{Teacher}
& \textbf{Student}
& \textbf{Fidelity ($R^2$)}
& \textbf{Decision Jaccard}
& \textbf{Spearman}
& \textbf{Kendall $\tau_b$}
& \textbf{Boundary Kendall $\tau_b$} \\

\midrule

Causal forest
& Decision tree
& 0.6272
& 0.7749
& 0.7663
& 0.6029
& -0.3686 \\

Causal forest
& Random forest
& 0.6923
& 0.7856
& 0.7813
& 0.6151
& -0.2872 \\

\addlinespace

BCF
& Decision tree
& 0.8503
& 0.9120
& 0.9316
& 0.8118
& 0.5985 \\

BCF
& Random forest
& 0.8512
& 0.9370
& 0.9468
& 0.8227
& 0.6301 \\

\bottomrule
\end{tabular}
}

\vspace{1mm}

\end{table*}

\begin{table*}[!h]
\centering

\caption{\textbf{NHANES smoking-history policy: Representative clinician-facing
examples from the causal-forest-distilled decision-tree and random-forest
students.} The shallow RF surrogate achieved an $R^2$ of $0.797$ with respect to the
full causal-forest-distilled RF predictions and is used only for
interpretation, not for policy allocation.}
\label{tab:smoking_cf_clinician_rules}

\resizebox{\textwidth}{!}{
\begin{tabular}{p{1.5cm}p{5.8cm}cccp{3.0cm}p{5.8cm}}
\toprule

\textbf{Student}
&
\textbf{Readable rule}
&
\textbf{Eligible $N$}
&
\textbf{Student benefit (pp)}
&
\textbf{CF teacher benefit (pp)}
&
\textbf{Policy status}
&
\textbf{Illustrative interpretation}
\\
\midrule

Decision tree
&
Age $\leq 60$ years, race/ethnicity other than non-Hispanic White,
and high-school education or less
&
245
&
8.33
&
7.87
&
Fully prioritized
&
This subgroup had comparatively high predicted benefit and was fully
prioritized under the $q=0.70$ intervention budget.
\\

Decision tree
&
Age $>60$ years, moderate-to-high income (PIR $\geq 2$),
and race/ethnicity other than non-Hispanic White
&
88
&
1.89
&
1.36
&
Capacity boundary
&
This subgroup had positive predicted benefit but lay near the intervention
capacity boundary, so only part of the subgroup was selected.
\\

Decision tree
&
Age $>60$ years, low income (PIR $<2$),
and high-school education
&
65
&
1.53
&
0.68
&
Outside capacity
&
Predicted benefit remained positive, but the subgroup was not prioritized
within the available intervention capacity.
\\

\addlinespace

Random forest Surrogate
&
Age $\leq 60$ years, race/ethnicity other than non-Hispanic White,
and high-school education or less
&
245
&
8.20
&
7.87
&
Fully prioritized
&
This surrogate region represents individuals with comparatively high
predicted benefit who were consistently prioritized by the RF policy.
\\

Random forest Surrogate
&
Age $\leq 60$ years, race/ethnicity other than non-Hispanic White,
and more than high-school education
&
229
&
2.26
&
2.00
&
Intermediate / mixed
&
This surrogate region contained individuals with positive benefit but
mixed allocation decisions around the policy boundary.
\\

Random forest Surrogate
&
Age $>60$ years, moderate-to-high income (PIR $\geq 2$),
and race/ethnicity other than non-Hispanic White
&
88
&
1.97
&
1.36
&
Outside capacity
&
Although the region showed positive predicted benefit, it fell outside the
allocation capacity of the RF policy.
\\

\bottomrule
\end{tabular}
}

\vspace{1mm}

\end{table*}

\begin{table*}[!h]
\centering

\caption{\textbf{NHANES smoking-history policy: Representative clinician-facing
examples from the BCF-distilled decision-tree and random-forest students.}
The shallow RF surrogate achieved an $R^2$ of $0.897$ with respect to the
full BCF-distilled RF predictions and is used only for interpretation, not
for policy allocation.}
\label{tab:smoking_bcf_clinician_rules}

\resizebox{\textwidth}{!}{
\begin{tabular}{p{1.5cm}p{5.8cm}cccp{3.0cm}p{5.8cm}}
\toprule

\textbf{Student}
&
\textbf{Readable rule}
&
\textbf{Eligible $N$}
&
\textbf{Student benefit (pp)}
&
\textbf{BCF teacher benefit (pp)}
&
\textbf{Policy status}
&
\textbf{Illustrative interpretation}
\\
\midrule

Decision tree
&
Age $<40$ years, at least high-school education, and race/ethnicity
other than Mexican American
&
274
&
6.07
&
6.04
&
Fully prioritized
&
This subgroup had comparatively high predicted benefit and was fully
prioritized under the $q=0.70$ intervention budget.
\\

Decision tree
&
Age 40--60 years, at least high-school education, and male
&
170
&
3.58
&
3.84
&
Capacity boundary
&
This subgroup had positive predicted benefit but lay near the intervention
capacity boundary, so only part of the subgroup was selected.
\\

Decision tree
&
Age $>60$ years, at least high-school education, and female
&
157
&
3.02
&
2.30
&
Outside capacity
&
Predicted benefit remained positive, but the subgroup was not prioritized
within the available intervention capacity.
\\

\addlinespace

Random forest
&
Age $<40$ years and at least high-school education
&
309
&
5.94
&
5.82
&
Fully prioritized
&
This surrogate region represents individuals with comparatively high
predicted benefit who were consistently prioritized by the RF policy.
\\

Random forest
&
Age 40--60 years and at least high-school education
&
373
&
4.38
&
4.43
&
Intermediate / mixed
&
This surrogate region contained individuals with positive benefit but
mixed allocation decisions near the policy boundary.
\\

Random forest
&
Age $>60$ years, at least high-school education, and male
&
182
&
1.52
&
1.11
&
Outside capacity
&
Although the region showed positive predicted benefit, it fell outside the
allocation capacity of the RF policy.
\\

\bottomrule
\end{tabular}
}

\vspace{1mm}
\end{table*}

\subsection{Ablation Study Results}


\subsubsection{Sensitivity to the outcome nuisance-model specification}

\begin{figure}[!h]
    \centering
\includegraphics[width=7cm,height=5cm]{ 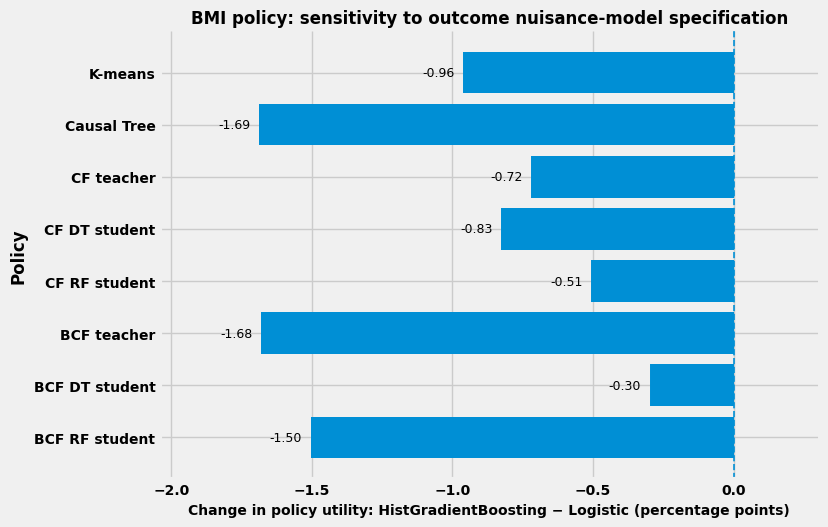}

    \caption{\textbf{BMI policy: sensitivity to the outcome nuisance-model
    specification.}
    Change in held-out policy utility after replacing the logistic-regression
    outcome nuisance model in the doubly robust evaluator with a
    HistGradientBoosting classifier. The dashed vertical line at zero indicates no change in
    estimated utility. Negative values indicate lower utility estimates under
    the HistGradientBoosting outcome nuisance model.Changes are shown in percentage points for visual readability.}

    \label{fig:bmi_histgrad_ablation}
\end{figure}

Figure~\ref{fig:bmi_histgrad_ablation} shows the change in held-out policy
utility when the logistic outcome nuisance regression in the DR evaluator was
replaced with a HistGradientBoosting classifier. Across the BMI policies,
the resulting changes were modest, ranging from $-0.30$ to $-1.69$ percentage
points. The mean absolute change in utility was $1.02$ percentage points and
the maximum absolute change was $1.69$ percentage points. HistGradientBoosting
yielded slightly lower utility estimates for all policies in this experiment.
Although the exact point-estimate ordering of several closely performing
policies changed, the overall magnitude and pattern of policy performance
remained broadly similar.

\begin{figure}[!h]
    \centering
    \includegraphics[width=7cm,height=5cm] { 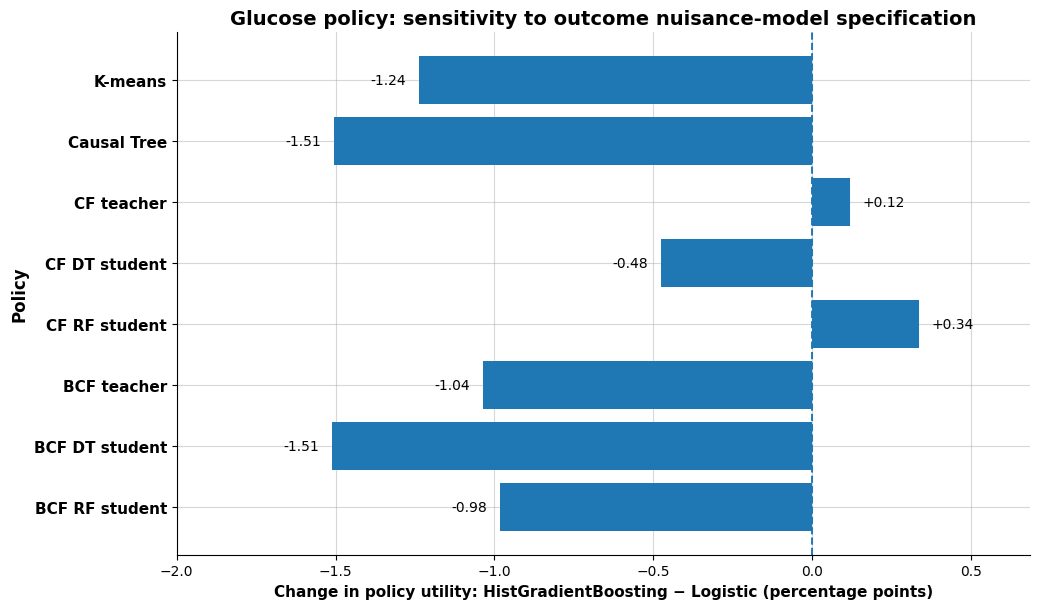}

    \caption{\textbf{Glucose policy: sensitivity to the outcome nuisance-model
    specification.}
    The dashed vertical line at zero indicates no change in
    estimated utility. Negative values indicate lower utility estimates under
    the HistGradientBoosting outcome nuisance model.Changes are shown in percentage points for visual readability.}
    \label{fig:glucose_histgrad_ablation}
\end{figure}

Figure~\ref{fig:glucose_histgrad_ablation} shows the change in held-out policy
utility when the logistic outcome nuisance regression in the DR evaluator was
replaced with a HistGradientBoosting classifier. Across the glucose policies,
the changes ranged from $-1.51$ to $+0.34$ percentage points. The mean absolute
change in utility was $0.90$ percentage points, and the maximum absolute change
was $1.51$ percentage points. HistGradientBoosting yielded slightly lower
utility estimates for most policies, while the CF teacher and CF random-forest
student showed small increases of $0.12$ and $0.34$ percentage points,
respectively. Although the exact point-estimate ranking changed among several
closely performing policies, the absolute changes in estimated utility were
small, with no policy changing by more than approximately $1.5$ percentage
points.

\begin{figure}[!h]
    \centering
  \includegraphics[width=7cm,height=5cm]{ 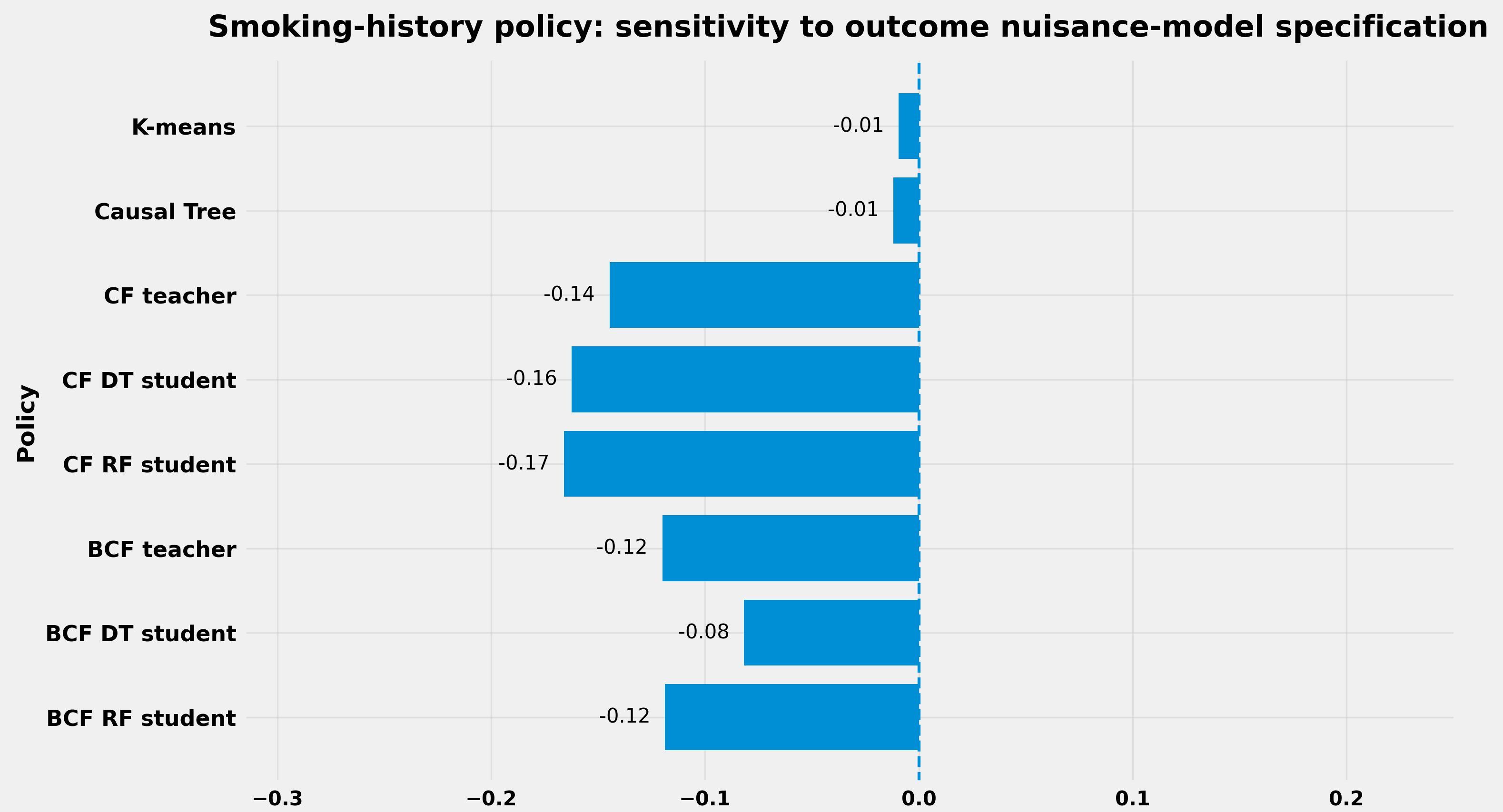}

    \caption{\textbf{Smoking policy: sensitivity to the outcome nuisance-model
    specification.}
    The dashed vertical line at zero indicates no change in
    estimated utility. Negative values indicate lower utility estimates under
    the HistGradientBoosting outcome nuisance model.Changes are shown in percentage points for visual readability.}
    \label{fig:smoking_histgrad_ablation}
\end{figure}

Figure~\ref{fig:smoking_histgrad_ablation} shows the change in held-out policy utility when the logistic outcome nuisance regression in the DR evaluator was replaced with a HistGradientBoosting classifier while all learned policies and other evaluation settings were held fixed. Across the smoking-history policies, the changes were uniformly small and negative, ranging from approximately $-0.17$ to $-0.01$ percentage points. Based on the displayed values, the mean absolute change in utility was approximately $0.10$ percentage points, and the maximum absolute change was approximately $0.17$ percentage points. The K-means and Causal Tree policies were essentially unchanged, while the largest decreases were observed for the CF random-forest student ($-0.17$ percentage points) and CF decision-tree student ($-0.16$ percentage points). Overall, replacing the logistic outcome nuisance model with HistGradientBoosting produced only minor downward shifts in estimated utility, indicating that the held-out smoking-policy evaluation was comparatively insensitive to this nuisance-model specification

\subsubsection{Sensitivity to the Discovery--Evaluation Split}

\begin{figure}[!h]
    \centering
    \includegraphics[width=7cm,height=5cm]{ 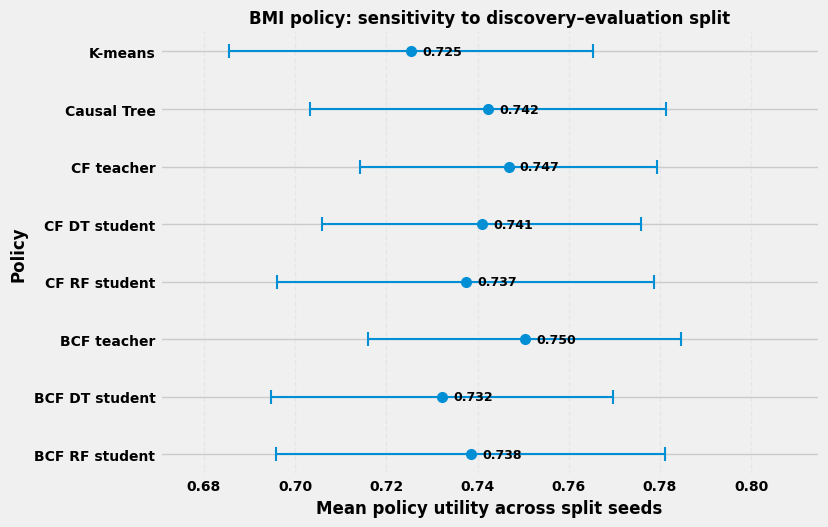}
    \caption{\textbf{BMI policy: sensitivity of held-out policy utility to the
    discovery--evaluation split.}
    Points show the mean held-out policy utility across the independent
    discovery--evaluation splits, and horizontal error bars show one standard
    deviation across splits. The BCF teacher achieved the
    highest mean utility point estimate followed closely by the
    CF teacher.}
    \label{fig:bmi_split_seed_utility}
\end{figure}

\begin{figure}[!h]
    \centering
    \includegraphics[width=7cm,height=5cm]{ 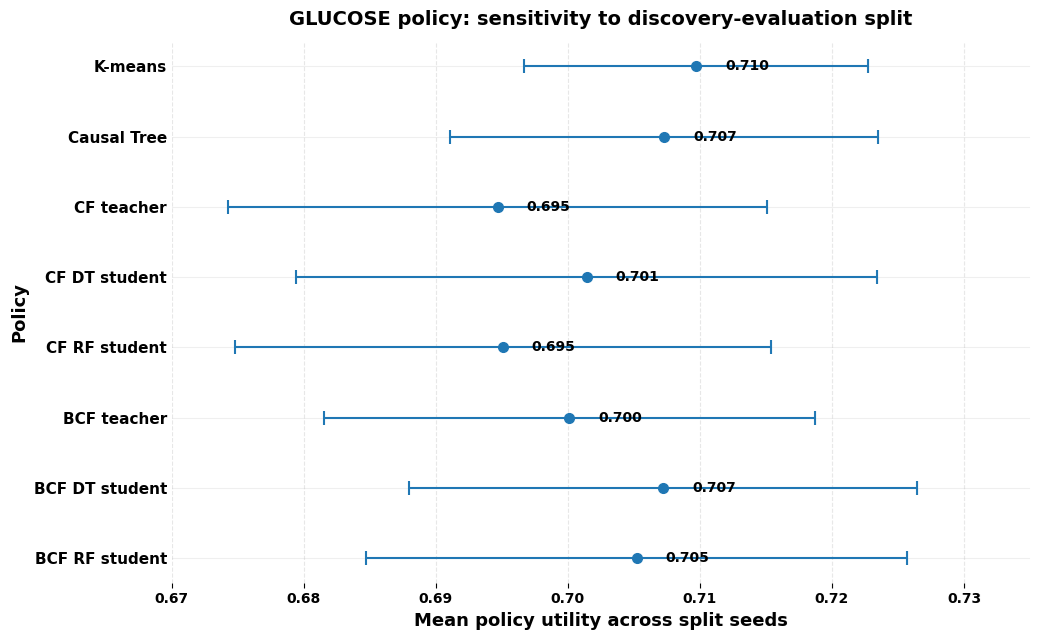}
\caption{\textbf{Glucose policy: sensitivity of held-out policy utility to the
discovery--evaluation split.}
 K-means attained the highest mean utility point estimate followed closely by the Causal Tree. Overall, the mean utilities were relatively similar across policies, with
overlapping split-to-split variability. Error bars summarize variability
across discovery--evaluation splits and are not confidence intervals.}
    \label{fig:glucose_split_seed_utility}
\end{figure}

\begin{figure}[!h]
    \centering
  \includegraphics[width=7cm,height=5cm]{ 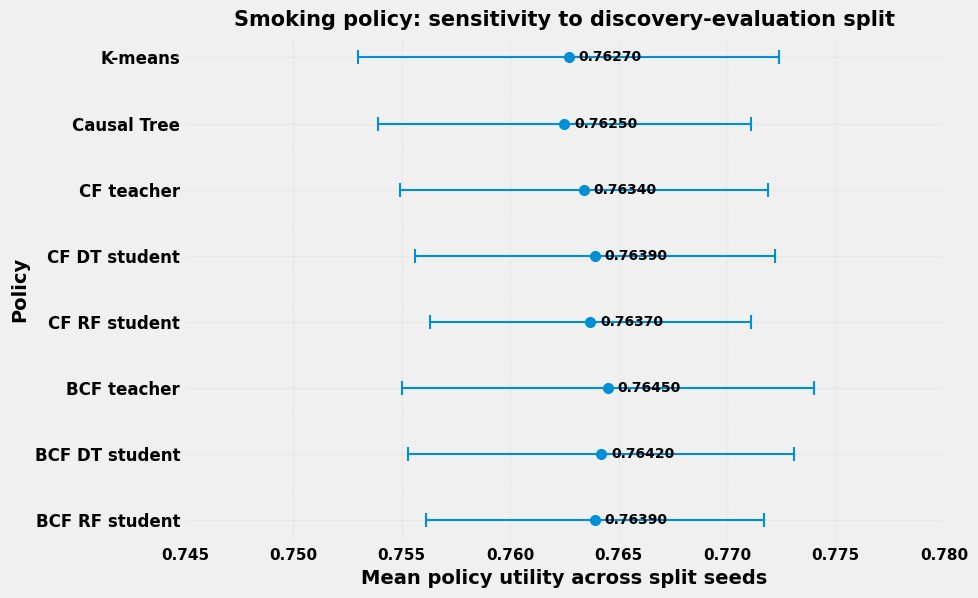}
    \caption{\textbf{Smoking policy: sensitivity of held-out policy utility to the
    discovery--evaluation split.}
     The BCF teacher attained the highest mean utility
    point estimate, followed closely by the BCF DT student. Overall, the policies showed closely
    comparable mean performance with substantial overlap in split-to-split
    variability. Error bars summarize variability across discovery--evaluation
    splits and are not confidence intervals.}
    \label{fig:smoking_split_seed_utility_causal}
\end{figure}

We assessed sensitivity to the discovery--evaluation partition across five independent random splits while holding all other analysis settings fixed. Mean held-out utilities were broadly similar across splits and qualitatively consistent with the representative split (Figs.~\ref{fig:bmi_split_seed_utility}--\ref{fig:smoking_split_seed_utility_causal}). The highest mean point estimate varied by application, with BCF highest for BMI and smoking-history and K-means highest for glucose.

These differences are descriptive comparisons of point estimates across the
examined splits. The standard-deviation bars quantify split-to-split
variability rather than inferential uncertainty.

\subsubsection{Sensitivity of policy utility to score-policy encoding}

\begin{figure}[!h]
    \centering
     \includegraphics[width=7cm,height=5cm]{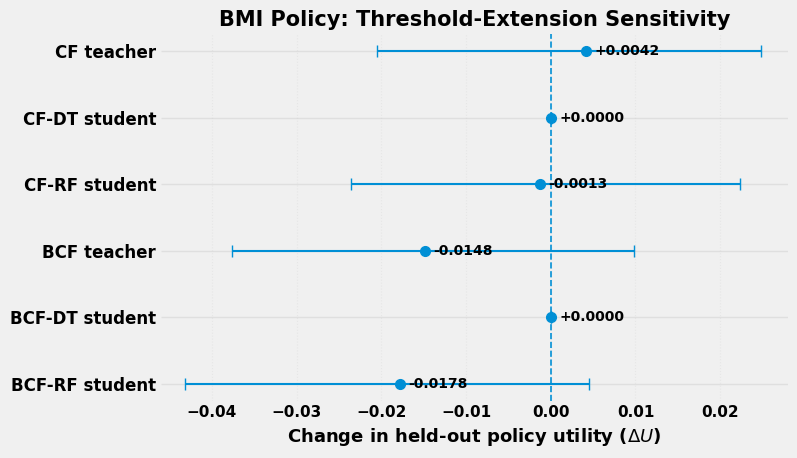}
    \caption{\textbf{BMI policy: sensitivity to score-policy threshold extension.} Horizontal bars denote 95\% bootstrap confidence intervals, and the dashed
    vertical line indicates no change in utility. Changes were small across all models }
    \label{fig:bmi_threshold_extension}
\end{figure}

\begin{figure}[!h]
    \centering
     \includegraphics[width=7cm,height=5cm]{ 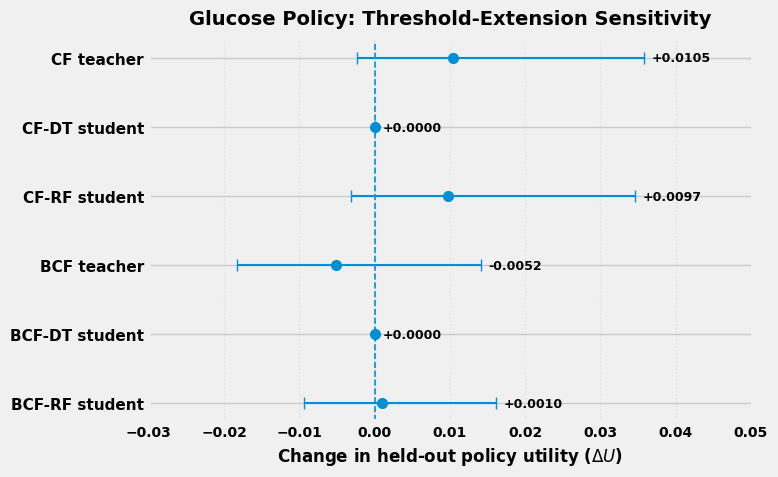}
    \caption{\textbf{Glucose policy: sensitivity to score-policy threshold extension.} Horizontal bars denote 95\% bootstrap confidence intervals, and the dashed
    vertical line indicates no change in utility. Changes were small across all models }
    \label{fig:glucose_threshold_extension}
\end{figure}

\begin{figure}[!h]
    \centering
    \includegraphics[width=7cm,height=5cm]{ 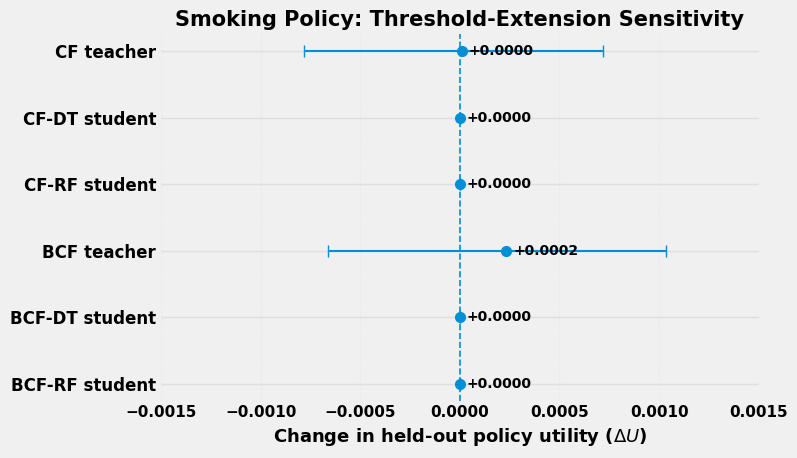}
    \caption{\textbf{Smoking policy: sensitivity to score-policy threshold extension.} Horizontal bars denote 95\% bootstrap confidence intervals, and the dashed
    vertical line indicates no change in utility. Changes were small across all models }
    \label{fig:smoking_threshold_extension}
\end{figure}

As shown in
Fig.~\ref{fig:bmi_threshold_extension}, this modification left the allocation
among the eligible individuals for the BMI policy unchanged for all six teacher and
student policies. Changes in held-out policy utility were small, with a mean
absolute change of 0.0064 and a maximum absolute change of 0.0178. The 95\%
bootstrap confidence intervals for the utility change included zero for all
policies, indicating that the BMI policy-value results were insensitive to
this alternative extension of the score policy to $T=0$ observations. As shown in Fig.~\ref{fig:glucose_threshold_extension}, extending the learned
score policy left the allocation among eligible individuals
unchanged for all six glucose teacher and student policies.  Changes in held-out
policy utility were small, with a mean absolute change of 0.0044 and a maximum
absolute change of 0.0105. As shown in Fig.~\ref{fig:smoking_threshold_extension}, extending the learned
score policy left the allocation among eligible individuals unchanged for all six
smoking teacher and student policies. Changes in held-out policy utility were
negligible, with a mean absolute change of 0.00004 and a maximum absolute
change of 0.00023.

\subsubsection{IHDP Dataset Results}

As the IHDP benchmark provides the ground-truth potential-outcome
functions, we evaluated policy performance using the true policy value rather than an estimated doubly robust value. Specifically, each method was used to rank individuals, and the corresponding policy was constructed under the common
70\% budget constraint. Its value was then computed from the known counterfactual outcomes.

While PEHE and policy value are related, they are not redundant. A formal decomposition (Supplementary Note 1, Section 0.2–0.3) shows that policy regret is jointly bounded by CATE-estimation error and the disagreement rate between the learned and oracle policies. We therefore report PEHE, true policy value, and oracle top-q Jaccard jointly rather than PEHE alone.

Across the 10 IHDP realizations, the BCF teacher demonstrated the strongest overall
performance among the teacher models. It achieved a lower mean
$\sqrt{\mathrm{PEHE}}$ and a higher mean true policy value than the causal
forest teacher (Table~\ref{tab:ihdp_mean_sd_performance}). This advantage was
also reflected in the ground-truth agreement measures. BCF attained higher
mean Spearman and Kendall correlations with the true CATE ranking and greater
overlap with the oracle top-$q$ policy than the causal forest teacher
(Table~\ref{tab:ihdp_fidelity_overlap}). Therefore, within this IHDP analysis, BCF
served as the more accurate teacher for both treatment-effect estimation and
for identifying individuals prioritized by the oracle budget-constrained
policy.

Among the distilled models, the BCF random-forest student demonstrated the best
overall performance. It attained the lowest mean $\sqrt{\mathrm{PEHE}}$ and the highest
mean true policy value among the student models, while also exhibiting the
strongest agreement with both its teacher and the known ground truth. In
particular, it achieved the highest teacher-ranking correlations, teacher
top-$q$ Jaccard, true-CATE rank correlations, and oracle top-$q$ Jaccard among
the distilled students. Within both teacher families, the random-forest
students generally preserved their teachers' rankings and selected populations
more closely than the corresponding decision-tree students.

The K-Prototypes results should be interpreted in light of the weak
cluster separation observed in the IHDP covariate space. Gower-based silhouette
coefficients were low across the prespecified candidate range, with the maximum attained at $K=3$. Thus, $K=3$ represented
the best-supported solution among the candidates considered, but the absolute
silhouette value did not indicate three sharply separated natural subgroups.
One contributing factor may be the composition of the IHDP covariate set,
which contains 19 binary and six continuous variables. It is likely that the substantial overlap in the
binary covariate profiles may have reduced the contrast between within-cluster
and between-cluster dissimilarities. More broadly, prior work has shown that pairwise-distance
contrast can deteriorate as dimensionality increases and that this behavior
depends on the selected distance metric
\cite{gower2,gower3}. In terms of performance, K-Prototypes achieved a slightly lower mean
$\sqrt{\mathrm{PEHE}}$ and a slightly higher mean true policy value, whereas
the causal tree attained somewhat higher correlations with the true CATE
ranking and greater overlap with the oracle top-$q$ population. These results
therefore do not indicate a clear overall advantage of supervised over
unsupervised subgroup construction. Rather, the single causal tree showed
somewhat better recovery of the oracle-prioritized subgroup, while
K-Prototypes showed slightly better average treatment-effect estimation and
policy-value performance. The causal-tree comparator consisted of one
interpretable, shallow tree rather than an ensemble.

\begin{table*}[!h]
\centering
\scriptsize

\caption{\textbf{IHDP semi-synthetic analysis:} Performance averaged across
10 IHDP realizations, reported as mean $\pm$ standard deviation where
available. True policy value denotes the average ground-truth outcome under
the semi-synthetic IHDP outcome-generating process if treatment were assigned
according to the policy, with higher values indicating better performance.
Policy regret is the difference between the oracle policy value and the value
achieved by the evaluated policy, with lower values indicating performance
closer to the oracle and zero denoting the oracle itself. The selective BCF policy achieved a higher true policy value than the unrestricted treat-all reference because budgeted targeting can avoid allocating treatment to individuals with low or negative treatment benefit. The absolute
magnitude of policy value does not have a direct clinical interpretation and
is used for comparative evaluation within the benchmark. The oracle
positive-benefit policy under the 70\% budget cap achieved a mean value of
$14.3667 \pm 16.4195$. PEHE is not applicable to the reference policies.
The comparatively large standard deviations reflect known variation in the
scale of the original IHDP outcome simulations across realizations
\cite{curth2021really}.}
\label{tab:ihdp_mean_sd_performance}

\resizebox{\textwidth}{!}{
\begin{tabular}{llccc}
\toprule
\textbf{Method}
& \textbf{Method family}
& \textbf{$\sqrt{\mathrm{PEHE}}$}
& \textbf{True policy value}
& \textbf{Mean policy regret} \\
\midrule

Causal Forest teacher
& Causal Forest
& $3.2190 \pm 5.4669$
& $14.0385 \pm 15.8533$
& $0.3282$ \\

CF Decision Tree student
& CF distillation
& $3.7957 \pm 6.5949$
& $13.8532 \pm 15.5185$
& $0.5135$ \\

CF Random Forest student
& CF distillation
& $3.5699 \pm 6.1307$
& $13.9602 \pm 15.7249$
& $0.4065$ \\

BCF teacher
& BCF
& $\mathbf{1.8410 \pm 3.4889}$
& $\mathbf{14.2940 \pm 16.3010}$
& $\mathbf{0.0727}$ \\

BCF Decision Tree student
& BCF distillation
& $3.2212 \pm 5.6655$
& $13.9773 \pm 15.7715$
& $0.3894$ \\

BCF Random Forest student
& BCF distillation
& $2.4207 \pm 4.4021$
& $14.1724 \pm 16.0666$
& $0.1943$ \\

K-Prototypes subgroup
& Unsupervised subgroup
& $4.0196 \pm 6.8354$
& $13.2687 \pm 15.0508$
& $1.0980$ \\

Causal Tree subgroup
& Supervised subgroup
& $4.4614 \pm 8.2784$
& $13.1436 \pm 13.8546$
& $1.2231$ \\

\midrule

Oracle positive-benefit budget cap
& Reference
& --
& $14.3667 \pm 16.4195$
& $0.0000$ \\

Treat none
& Reference
& --
& $9.4531 \pm 12.2658$
& $4.9136$ \\

Treat all (unrestricted reference)
& Reference
& --
& $14.0893 \pm 13.8702$
& $0.2774$ \\

\bottomrule
\end{tabular}
}
\end{table*}

\begin{table*}[!h]
\centering

\caption{\textbf{IHDP ranking fidelity and policy overlap:}
Results are averaged across 10 IHDP realizations and reported as mean
$\pm$ standard deviation. Spearman and Kendall coefficients measure agreement
in the complete treatment-benefit ranking, whereas top-$q$ Jaccard measures
overlap between the populations selected under the 70\% treatment budget.
Teacher-agreement measures are reported only for distilled students.}
\label{tab:ihdp_fidelity_overlap}
\resizebox{\textwidth}{!}{
\begin{tabular}{llcccccc}
\toprule
&
&
\multicolumn{3}{c}{\textbf{Agreement with teacher}}
&
\multicolumn{3}{c}{\textbf{Agreement with ground truth}} \\
\cmidrule(lr){3-5}
\cmidrule(lr){6-8}
\textbf{Method}
& \textbf{Teacher}
& \textbf{Spearman}
& \textbf{Kendall}
& \textbf{Top-$q$ Jaccard}
& \textbf{Spearman}
& \textbf{Kendall}
& \textbf{Oracle top-$q$ Jaccard} \\
\midrule

Causal Forest teacher
& --
& --
& --
& --
& $0.7607 \pm 0.1310$
& $0.5760 \pm 0.1190$
& $0.7681 \pm 0.0526$ \\

CF Decision Tree student
& Causal Forest
& $0.8172 \pm 0.1189$
& $0.6775 \pm 0.1332$
& $0.8112 \pm 0.0788$
& $0.6694 \pm 0.1882$
& $0.5283 \pm 0.1598$
& $0.7361 \pm 0.0754$ \\

CF Random Forest student
& Causal Forest
& $0.9001 \pm 0.0719$
& $0.7432 \pm 0.1018$
& $0.8621 \pm 0.0590$
& $0.7308 \pm 0.1353$
& $0.5477 \pm 0.1171$
& $0.7555 \pm 0.0361$ \\

\midrule

BCF teacher
& --
& --
& --
& --
& $\mathbf{0.8572 \pm 0.1573}$
& $\mathbf{0.7016 \pm 0.1643}$
& $\mathbf{0.8601 \pm 0.0548}$ \\

BCF Decision Tree student
& BCF
& $0.7864 \pm 0.0710$
& $0.6359 \pm 0.0824$
& $0.8052 \pm 0.0625$
& $0.7252 \pm 0.1312$
& $0.5792 \pm 0.1307$
& $0.7742 \pm 0.0402$ \\

BCF Random Forest student
& BCF
& $\mathbf{0.9236 \pm 0.0379}$
& $\mathbf{0.7767 \pm 0.0636}$
& $\mathbf{0.8778 \pm 0.0417}$
& $\mathbf{0.8261 \pm 0.1353}$
& $\mathbf{0.6577 \pm 0.1322}$
& $\mathbf{0.8153 \pm 0.0452}$ \\

\midrule

K-Prototypes subgroup
& --
& --
& --
& --
& $0.4070 \pm 0.2602$
& $0.3204 \pm 0.2081$
& $0.5421 \pm 0.0713$ \\

Causal Tree subgroup
& --
& --
& --
& --
& $0.4269 \pm 0.2550$
& $0.3380 \pm 0.2067$
& $0.6283 \pm 0.0822$ \\

\bottomrule
\end{tabular}
}
\end{table*}

\section{Discussion}
\label{discussion_causal}

The primary finding is that no single policy representation consistently outperformed others across all applications. Various approaches achieved comparable aggregate policy values, yet differed in structural complexity, allocation granularity, and fidelity near the intervention boundary. Therefore, policy representations should be assessed not only by average utility but also by the specific decisions they generate.

The BMI experiment demonstrated most clearly that a flexible causal teacher model can be compressed without substantial loss in policy performance. The causal-tree comparator achieved the highest point-estimate utility among budget-constrained policies, closely followed by the BCF teacher, the causal forest teacher, and several distilled student models. The principal conclusion from this experiment is not a statistically significant ranking of methods, but rather that both relatively simple subgroup policies and distilled student models can attain policy performance comparable to that of more complex causal models.

The clinician-facing rules further indicated that certain qualitative aspects of treatment-effect structure remained stable across different teacher models, despite variations in exact
data-adaptive thresholds. Specifically, elevated
DPF values were consistently observed in higher-benefit regions identified by both CF and BCF teachers. Both random-forest student models identified highly prioritized profiles characterized by relatively few pregnancies and elevated familial-risk values. However, the exact split locations and secondary modifiers varied between teacher models.

The glucose experiment yielded a different ranking of methods. In this case, the shallow
causal-tree subgroup comparator attained the highest held-out point-estimate
utility, closely followed by K-means and the CF-distilled decision tree, while the remaining teacher and student policies also demonstrated similar performance. This finding cautions against assuming that increased model
flexibility necessarily results in higher policy value across all applications.
When treatment-effect heterogeneity can be adequately captured by a small
number of partitions, a simple subgroup rule may perform comparably to an
individualized ensemble model.

Nevertheless, the glucose rules revealed meaningful structural similarities across teacher models.
Moderately high to high BMI values were consistently present in higher-priority
regions identified by both CF and BCF distillation, whereas the role of
DPF was less stable and depended more strongly on the
teacher and the accompanying covariate partitions. Thus, the glucose
experiment further illustrates the importance of comparing
across plausible teacher models rather than interpreting a single fitted tree
in isolation. Recurrent qualitative features provide stronger descriptive
evidence than reliance on any single numerical threshold.

The smoking-history experiment provides perhaps the clearest example of the
distinction between aggregate policy value and individual allocation.
K-means obtained the highest held-out point-estimate utility among the
budget-constrained policies, but the causal forest, BCF family, causal tree,
and distilled students achieved closely related values. 
Therefore, aggregate utility alone would suggest relatively
little distinction among the approaches.

The allocation-disagreement analysis gives a different view. K-means and the
causal-forest policy used the same expected intervention capacity but agreed
on only part of the selected population. The individuals selected by one
policy but not the other also differed systematically in their covariate
profiles and CF-estimated benefits. In particular, the K-means-only group had
negative mean benefit under the CF scoring function, whereas the CF-only group
had positive mean CF-estimated benefit, as shown in the supplementary files. Since these contrasts are evaluated
using the CF teacher's own effect estimates, they are not independent evidence
that the CF allocation is causally superior.

The smoking-history clinician-facing rules further showed both stability and
teacher dependence. Age emerged repeatedly under both CF and BCF, with younger
or middle-aged groups tending to occur in more highly prioritized regions and
older groups appearing more frequently near or outside the allocation
boundary.

Taken together, the subgroup experiments do not support a simple conclusion
that supervised subgroup construction is uniformly superior to unsupervised
clustering. The causal tree could isolate sharper high- and negative-benefit
regions in some experiments because treatment-effect information informs its
partitioning, whereas K-means constructs groups solely from pretreatment
covariate similarity. However, this additional supervision can also create
small leaves with unstable effect estimates, and the causal tree did not
consistently achieve higher held-out utility than K-means. More broadly,
the subgroup policies considered here rank clusters or leaves using
point-estimated mean benefit and aggregate gain without incorporating
estimation uncertainty into subgroup admission. In settings where an
incorrect allocation may carry substantial clinical cost, a favorable point
estimate alone may provide insufficient evidence for intervention,
particularly for small or heterogeneous subgroups. Uncertainty-aware
admission criteria, such as lower-confidence-bound or posterior-probability
rules \cite{acharya2026unsupervised} therefore represent an important extension for higher-risk applications.

The more important distinction is therefore between \emph{group-level} and
\emph{individualized} resolution. Unsupervised clustering has an important
practical advantage. It can generate a small number of recognizable
pretreatment profiles without using outcome or estimated treatment-effect
information to construct those groups. Such policies can be easy to describe,
audit, and implement. However, the disagreement analyses demonstrate what this
simplicity can cost. Individuals assigned to the same subgroup receive the
same or a common fractional allocation even when an individualized causal
model places them at different positions in the benefit ranking. Consequently, similar aggregate utility does not imply similar allocation or equivalent
clinical interpretation. Another important distinction concerns how learned policies are transferred to the evaluation cohort. In the cases of K-means and the Causal Tree, subgroup definitions and subgroup priority ordering remain fixed based on the discovery cohort. In contrast, for the teacher and distilled-student policies, the learned benefit-scoring function is fixed and subsequently applied to the evaluation cohort to rank eligible individuals according to the budget constraint.

The teacher-student analyses indicate a practical middle ground between these
two extremes. A flexible causal teacher can be used offline to estimate a
confounding-adjusted treatment-effect surface, after which a simpler student
can approximate the treatment ranking used for deployment. The decision-tree
students provide the strongest structural simplification and explicit
root-to-leaf rules, whereas random-forest students retain greater flexibility
at the cost of reduced direct interpretability. The appropriate student
therefore depends on the desired balance between transparency and fidelity
rather than on predictive $R^2$ alone.

The structural-complexity results also require some attention. Compression was
particularly large relative to the causal-forest teachers, whose ensembles
contained many more terminal leaves than the students. In contrast, the BCF
treatment-effect forest is itself strongly regularized and can have relatively
few terminal leaves. Consequently, an RF student is not guaranteed to contain
fewer raw leaves than the BCF treatment-effect component. Its deployment
advantage instead arises from replacing a posterior distribution over causal
forests with a single fixed prediction rule. Thus, terminal-leaf count is one
useful measure of structural complexity, but should be interpreted together
with whether inference requires an ensemble, posterior sampling, and
additional prognostic components.

A key implication of the distillation results is that global
teacher--student agreement is not sufficient to determine whether a student
can operationally replace its teacher. Policy decisions under a fixed budget
are determined by relative ordering near the allocation cutoff. A student can
therefore achieve a high global $R^2$, Spearman correlation, or Kendall
coefficient while reversing the order of individuals immediately above and
below the operational threshold. Such errors may have little influence on
global fidelity statistics but directly change who receives the intervention.

For this reason, boundary-restricted Kendall agreement should be considered a
decision-focused diagnostic rather than an ancillary prediction metric.
The BMI experiment demonstrated weak or negative local ordering for several student models
despite substantially stronger global rank agreement.
The smoking experiment provided an especially informative contrast.
The BCF-distilled student models preserved positive local ordering near the boundary, whereas the CF-distilled
student models exhibited negative boundary Kendall coefficients despite positive
global correlations.
These results suggest that replacing a causal teacher with a student model based solely on global predictive fidelity is not advisable.
Approximation should instead demonstrate acceptable policy value, selected-set
agreement, global ranking fidelity, and local ordering around the intended
operating threshold.
When local fidelity is poor, it is more defensible to retain the teacher model for borderline cases \cite{rawat2021doubt} or to report the student model as an approximation rather than as an
interchangeable replacement.

The paired-bootstrap analyses reinforce this interpretation. Point-estimate rankings differed across the BMI, glucose, and smoking-history experiments, but the prespecified comparisons did not remain statistically significant after Holm adjustment. Therefore, the observed ordering of learned policies should be interpreted descriptively rather than as evidence that the highest-ranked method is generally superior. Importantly, lack of a significant paired difference does not establish equivalence \cite{piaggio2006reporting}. Instead, the results may reflect both genuinely small differences in aggregate policy value and limited precision. Modest held-out sample sizes and variability in doubly robust estimation limit the resolution of small differences. These findings make decision-focused properties, including interpretability, allocation overlap, boundary fidelity, and structural complexity, particularly relevant when several policies achieve comparable estimated value.

The sensitivity analyses broadly supported the stability of the observational results. Replacing the logistic-regression outcome nuisance model with HistGradientBoosting produced only modest changes in held-out utility, with the smallest changes observed for the smoking-history policy. Similarly, across five discovery and evaluation splits, relative performance remained broadly consistent, although the highest mean-utility method differed across BMI, glucose, and smoking. These findings indicate that the overall conclusions are not driven by a particular nuisance-model specification or a single data split. Small differences in point-estimate utility should be interpreted descriptively rather than as stable evidence of superiority. The score-policy encoding sensitivity analysis provided additional reassurance. Extending the learned score thresholds to $T=0$ observations while preserving
the eligible $T=1$ allocation produced only small utility changes for BMI and
glucose and negligible changes for smoking. It thus
supported the robustness of the policy-value conclusions to this alternative
score-policy representation. 

The semi-synthetic IHDP experiments provide an important complement to the
observational analyses because true potential outcomes and treatment effects
are available. Across the ten realizations, BCF achieved the strongest
combination of treatment-effect estimation accuracy, true policy value, and agreement with the oracle allocation among the teachers. Among the distilled
models, the BCF random-forest student model provided the strongest overall
performance and more closely preserved both teacher and oracle rankings than
the corresponding decision-tree student models. These findings provide ground-truth evidence that the teacher-student strategy can retain much of
the decision-relevant information in a high-quality causal teacher model, rather
than merely reproducing its estimated value under an observational evaluator.

Another way to compare is to use a random allocation that fits the budget.
The expected value of randomly treating 70\% of individuals is the same as the value when no one is treated plus 0.70 times the difference between treating everyone and treating no one, which is about $12.70$ when calculated using the average values in Table~\ref{tab:ihdp_mean_sd_performance} \cite{imai}. All of the learned policies did better than this value on average. This shows that the CATE-based and subgroup-based rankings were able to find real differences in benefit that a random allocation without covariates would miss. This comparison uses averages from different runs. A paired comparison for each run, using
each policy's actual proportion treated, still needs to be done.

The IHDP experiments also clarify why treatment-effect prediction accuracy should not be
used as the sole criterion for policy selection.
As derived in the Supplementary Material, policy regret depends jointly on treatment-effect
estimation error and disagreement between the learned and oracle policies.
Thus, PEHE measures the numerical accuracy of the estimated treatment effects,
whereas policy value, policy regret, ranking agreement, and top-$q$ overlap
measure consequences for the downstream allocation.
Two estimators can have different PEHE yet induce similar treatment rankings, while comparatively
small numerical errors can have substantial impact when they occur around the budget
boundary.
This theoretical relationship is consistent with the empirical teacher-student analyses and provides an additional rationale for reporting
decision-focused fidelity measures alongside conventional treatment-effect
prediction metrics.

The results therefore favor neither purely supervised nor purely unsupervised
policy construction in all settings.
If the principal requirement is a small, highly transparent set of population-level rules, pretreatment subgroup
policies remain useful and can attain competitive average utility.
A supervised causal tree provides an alternative when effect-informed subgroup
construction is acceptable, although small-leaf instability must be monitored.
When the primary objective is to preserve individual treatment-effect heterogeneity and allocate a constrained intervention as efficiently as
possible, the IHDP results favor first estimating
effects with a flexible causal teacher model and then considering distillation for
deployment.

Within that strategy, the simplest student model should not automatically be
selected. The deployment model should be chosen using a hierarchy of
criteria including held-out policy value and uncertainty, teacher-student allocation
overlap, global ranking agreement, local fidelity around the operational
boundary, and finally structural and rule complexity.
A shallow decision tree is preferable when transparent rules are paramount and these fidelity
criteria remain acceptable. A small random forest is preferable when modest
additional complexity produces materially better preservation of teacher
ranking or allocation. If neither student model preserves the teacher model near the
decision boundary, the teacher model should not be regarded as fully replaceable.

\section{Limitations and Future Work}\label{limit_future}

This study is subject to several important limitations. The primary analyses rely on observational data, so causal relationships and heterogeneous treatment-effect estimates should be interpreted as conditional on the assumed causal structure and identification assumptions, rather than as definitive causal evidence. Identification depends on standard assumptions, including unconfoundedness (ignorability), consistency, and positivity. Violations of these assumptions, such as residual or unmeasured confounding or insufficient treatment overlap, may bias both the estimated CATEs and subsequent policy evaluations. Dataset-specific limitations are also present. In the NHANES analyses, each respondent is treated as a single observational unit, and survey weights are not incorporated to produce population-representative point estimates. The uncertainty quantification does not fully account for the complexity of the NHANES sampling design, including stratification and clustering. Therefore, the reported estimates and confidence intervals should not be interpreted as fully design-based, nationally representative NHANES estimates. The IHDP analysis is limited by reduced empirical overlap in certain regions of the covariate space under the standard de-randomized benchmark construction. As a result, IHDP is used primarily to evaluate CATE estimation, policy ranking, and ground-truth policy value, rather than to claim that full-population CATEs would necessarily be identifiable in a real observational study with similarly limited overlap. The policy experiments should be interpreted as hypothetical state interventions, not as evaluations of a known historical decision policy. For modifiable risk-factor exposures such as obesity and elevated glucose, the policies represent counterfactual changes in exposure state among prioritized eligible individuals and do not correspond to an implemented clinical program or bandit logging policy. The NHANES smoking exposure presents an additional limitation because it is defined using an ever-smoking criterion, which does not distinguish current from former smokers. The corresponding intervention should therefore be interpreted as a hypothetical change in smoking-history state, not as an acute smoking-cessation intervention among current smokers. Mixing current and former smokers within the exposed group may attenuate estimated effects and policy-value differences, limiting the clinical specificity of this intervention. Additional limitations relate to subgroup construction and subgroup-level effect estimation. K-means clustering produces hard cluster assignments and does not quantify uncertainty in cluster membership. Individuals near cluster boundaries may therefore be assigned somewhat arbitrarily under the primary hard-clustering analysis. Finally, the distilled student models are supervised regressors trained to approximate CATE surfaces estimated by the causal-forest and BCF teachers. The decision-tree and random-forest students do not estimate causal effects independently or optimize a causal objective. Their causal interpretation is inherited from the corresponding teacher and may therefore inherit biases arising from teacher misspecification.

Despite good empirical performance across multiple policies, our study identifies several directions to strengthen causal validity, improve deployability, and broaden applicability beyond the current binary state-intervention setting. Our distillation experiments focus on two interpretable student families, decision trees and random forests, and a broader exploration of the teacher--student design space remains open. Future work could also consider uncertainty-aware distillation targets, such as the posterior standard deviation of the CATE, a lower credible bound, or the posterior probability of benefit. Such targets would allow student models to approximate not only the teacher's mean effect surface but also aspects of its uncertainty-aware decision structure. As an additional effect-discretization baseline, we plan to apply the method presented in \cite{PCM} to partition the teacher's CATE surface into a small number of effect levels and evaluate the policies induced by these discretized representations. Several extensions could also strengthen the empirical design. The causal structures and treatment-effect patterns examined here are based primarily on cross-sectional data. An important next step is to evaluate whether the same relationships and patterns persist in longitudinal datasets, where temporal ordering can be represented more explicitly, and time-varying exposures and outcomes can be studied. Our current handling of missing data is based on complete-case analysis, which can reduce the effective sample size and may introduce selection bias when missingness is not completely random. Future work could therefore investigate causal-structure-informed imputation strategies that use prespecified causal relationships to guide imputation while avoiding inappropriate conditioning or the introduction of spurious dependencies. For NHANES specifically, future analyses will incorporate survey-statistical variance estimators that explicitly account for strata and primary sampling units, together with appropriate survey weights where population-representative inference is desired. Finally, we plan to extend the framework to additional clinically motivated intervention settings. One example is antiseizure-medication (ASM) optimization, motivated by evidence that ASMs can alter sleep architecture and may contribute to sleep disturbance in some settings \cite{asm_effects1,asm_effects2,asm_effects3,asm_effects4}. More broadly, future work could extend the framework beyond binary exposure-state modifications to include multivalued, continuous, or longitudinal intervention policies, thereby allowing a wider range of clinically relevant treatment and risk-factor optimization problems to be considered.

\section{Conclusion} \label{conclusion_causal}

Overall, the experiments suggest that interpretable causal policy design is
best viewed as a tradeoff among causal estimation, allocation resolution, and deployment complexity rather than as a contest between a single supervised
and unsupervised algorithm.
Unsupervised subgroup policies offer transparent population-level decisions but can mask within-group heterogeneity.
Causal trees provide effect-informed but still coarse partitions.
Flexible causal teacher models provide individualized rankings but are more difficult to deploy.
Distillation can recover much of their decision structure in simpler
models.
The most consequential approximation errors are those that change allocation near the operating threshold, motivating evaluation of policy
agreement and local ranking fidelity in addition to conventional prediction
metrics.
Under this view, the role of the causal teacher model is to estimate the heterogeneous intervention signal, while the role of the student model is to
provide the simplest deployment representation that preserves the
decision-relevant features of that signal.

\section*{Acknowledgements}
The authors express their gratitude to Madeline C. Fields and Lara V. Marcuse of the Icahn School of Medicine at Mount Sinai and the Mount Sinai Hospital, New York, NY, USA, for their assistance in elucidating potential causal relationships relevant to this study. Appreciation is also extended to Dr. Vishwas D. Pai of Pai Onco Care Center and Hubli Scan Center, India, for his valuable guidance and support.

\bibliography{plos_bibtex_sample}

\end{document}